\documentclass{article}
\usepackage{iclr2026_conference,times}

\usepackage{hyperref}
\usepackage{url}
\usepackage{graphicx}
\usepackage{algorithm}
\usepackage{algorithmic}
\usepackage{amsmath,amsfonts,bm,amssymb}
\usepackage{amsthm}
\usepackage{booktabs}
\usepackage{multirow}
\usepackage{colortbl}
\usepackage{subcaption}
\usepackage{xcolor}
\usepackage{enumitem}
\usepackage{wrapfig}
\usepackage[capitalize]{cleveref}
\Crefname{figure}{Fig.}{Figs.}
\Crefname{table}{Tab.}{Tabs.}
\Crefname{section}{Sec.}{Secs.}
\Crefname{algorithm}{Alg.}{Algs.}

\def\method{TASA}
\graphicspath{{figures/}{fig_materials/}}

\DeclareMathOperator*{\argmax}{arg\,max}

\DeclareMathOperator{\tr}{tr}

\newtheorem{theorem}{Theorem}
\newtheorem{definition}{Definition}
\newtheorem{proposition}{Proposition}
\newtheorem{remark}{Remark}

\definecolor{tasablue}{RGB}{31,119,180}
\definecolor{tasaorange}{RGB}{255,127,14}

\title{Beyond Activation Alignment:\\The Alignment-Diversity Tradeoff in Task-Aware LLM Quantization}

\author{
Fei~Wang$^{1}$ \quad
Chao~Xue$^{2}$ \quad
Taoran~Liu$^{4}$ \quad
Li~Shen$^{3,5}$ \quad
Ye~Liu$^{1}$ \quad
Changxing~Ding$^{1}$\thanks{Corresponding author.} \\
\\
{\small $^{1}$South China University of Technology \quad
$^{2}$JD.com} \\
{\small $^{3}$Shenzhen Campus of Sun Yat-sen University \quad
$^{4}$Ocean University of China} \\
{\small $^{5}$Center for AI Theoretical Foundation and Systems, Shenzhen Loop Area Institute} \\
\\
{\small \texttt{ft\_feiw@mail.scut.edu.cn, chxding@scut.edu.cn}}
}

\iclrfinalcopy

\begin{document}

\maketitle

\begin{abstract}
Mixed-precision quantization (MPQ) has become a key technique for deploying large language models under stringent memory and compute constraints.
We first identify a phenomenon that we term the \emph{Perplexity Illusion}: layers ranked as important by perplexity-based sensitivity show little rank correlation with those that are most influential for complex reasoning performance, with Kendall $\tau \approx 0$ in our analysis.
We further reveal an \emph{Alignment-Diversity Tradeoff}: using only target-task calibration data can degrade post-quantization performance, whereas incorporating general-domain data stabilizes sensitivity estimation and improves robustness across tasks.
Based on these observations, we propose \textbf{\method{}} (\textbf{T}ask-\textbf{A}ware \textbf{S}ensitivity \textbf{A}nalysis), a two-level framework that jointly optimizes calibration-data composition and mixed-precision bit allocation.
Specifically, \method{} searches for a calibration-data mixture using a training-free gradient-trace alignment criterion, and then aggregates perplexity and reasoning-oriented sensitivity signals to guide both inter-layer and intra-layer bit allocation.
Experiments on LLaMA-3-8B and Qwen2.5-7B reveal a \emph{precision inversion}: appropriately allocated 3.5-bit models can match or surpass less task-aware 4-bit baselines.
At an average precision of 3.5 bits, \method{} matches or outperforms several competitive 4-bit uniform baselines in aggregate accuracy, and improves over the strongest W3 baseline on GSM8K by more than 20 absolute points on LLaMA-3-8B. These results show that calibration-data composition substantially affects task-sensitive quantization, a factor underexplored in prior work.\footnote{Code is available at \url{https://github.com/WangFei-2019/TASA}.}
\end{abstract}

\section{Introduction}
\label{sec:intro}

Large language models (LLMs) have achieved strong performance across complex domains such as mathematical reasoning, scientific question answering, and code generation \citep{grattafiori2024llama,yang2024qwen2,deepseek2025r1,cobbe2021gsm8k,lewkowycz2022minerva,clark2018arc,chen2021codex}. However, their large memory footprint remains a barrier to deployment, particularly on resource-constrained edge devices and local servers. Post-training quantization (PTQ), which compresses full-precision weights into low-bit integers without retraining, has become the standard approach \citep{frantar2023optq,lin2024awq,kim2025guidedquant,he2025cola,li2026osaq,xu2026smoothness}. Among various PTQ paradigms, mixed-precision quantization (MPQ) further optimizes the compression--quality tradeoff by allocating higher bit-widths to sensitive layers while aggressively compressing the rest \citep{dong2019hawq,huang2024slim,zhao2025coopq,deng2026gemq}, enabling sub-4-bit average precision with minor quality loss.

The conventional MPQ pipeline follows a standardized recipe inherited from early quantization work. It collects a generic calibration dataset (e.g., WikiText-2) \citep{ji2025beware,li2025selfcalib}, profiles per-layer sensitivity using perplexity (PPL) or Hessian-based metrics on this data \citep{fang2025perplexity}, and solves bit-width allocation via greedy or dynamic programming search \citep{dong2019hawq,dong2020hawqv2,tang2024aptq}. This generalized workflow implicitly encodes two assumptions: (i)~\emph{perplexity serves as a faithful proxy for all downstream dimensions of model capability}, and (ii)~\emph{generic text distributions provide a sufficient calibration manifold for any targeted task}. While these assumptions held for small-scale language models, they have not been revisited for modern LLMs \citep{li2025quantmeetreason,liu2025quanthurts,zhou2026failuremodes}, where user demands have shifted from fluent next-token completion to multi-step logical inference.

In this paper, we challenge both assumptions and show that a model optimized to predict Wikipedia tokens does not necessarily excel at complex reasoning. Through systematic layer-wise sensitivity profiling under quantization perturbations, we reveal a disconnect between perplexity sensitivity and reasoning sensitivity. The ordinal correlation between PPL-sensitive layers and reasoning-sensitive layers is statistically indistinguishable from zero, with Kendall $\tau \approx 0$. While PPL sensitivity is concentrated at the input embeddings and output projections, the layers critical for mathematical and scientific reasoning lie in the middle block of the network, which supports multi-step computation and compositional inference. We formalize this phenomenon as the \textbf{Perplexity Illusion}. This implies that conventional PPL-guided allocation under-protects reasoning-critical layers, systematically safeguarding the wrong parameters for specialized deployments.

A natural question follows: if task alignment matters most, should we calibrate exclusively with task-specific data? The answer, however, is no. Calibrating solely with target-task distributions such as pure mathematical text consistently degrades quantization quality, despite maximizing activation alignment with the target domain. We term this the \textbf{Alignment-Diversity Tradeoff}: task-specific data offers the necessary activation alignment to identify task-relevant weight vectors, but general-domain data provides distributional diversity that regularizes the calibration Hessian and protects the model's generalization manifold from representation collapse. The global optimum lies at an intermediate mixing ratio that varies across models and target tasks, ruling out fixed heuristic recipes.

To jointly address calibration composition and bit-width allocation, we propose \textbf{\method{}} (\textbf{T}ask-\textbf{A}ware \textbf{S}ensitivity \textbf{A}nalysis), a two-level mixed-precision framework.
\method{} operates along two complementary axes. On the \emph{data axis}, it employs a training-free gradient trace alignment search to determine the optimal mixing ratio between general and task-specific calibration data, requiring only $(|\mathcal{A}|+1)$ forward passes. On the \emph{network axis}, \method{} introduces a Multi-Objective Aggregation (MOA) metric that fuses perplexity and reasoning signals into a single composite objective, formulating the constrained inter-layer allocation as a sequential resource-allocation problem (an ILP with knapsack structure) that admits an exact dynamic-programming solution. Layer-level precision is further refined via second-order salience-driven group-wise assignment at the intra-layer level.

Experiments across eight benchmarks spanning reasoning, commonsense, and language modeling show that \method{} advances the Pareto frontier for low-bit LLM compression. On LLaMA-3-8B, \method{} at 3.5-bit average precision matches several 4-bit uniform baselines while using 12.5\% fewer bits, retaining 97.2\% of FP16 aggregate accuracy. On Qwen2.5-7B, a 3.75-bit \method{} retains 99.1\% of FP16 performance. This \emph{precision inversion} is accompanied by large gains in mathematical reasoning, with the gap over uniform W3 baselines exceeding 20 points on GSM8K.

Our contributions:
\begin{itemize}
    \item \textbf{A Unified Tradeoff.} We formalize the \emph{Alignment-Diversity Tradeoff} in post-training quantization calibration and reveal the \emph{Perplexity Illusion}, showing that PPL sensitivity is statistically uncorrelated with reasoning sensitivity.
    \item \textbf{A Joint Optimization Framework.} \method{} automatically optimizes calibration composition via gradient trace alignment, aggregates multi-objective sensitivity signals, and solves joint inter- and intra-layer bit allocation. 
    \item \textbf{Strong Empirical Results.} On LLaMA-3-8B and Qwen2.5-7B, \method{} at 3.5-bit matches several 4-bit uniform baselines in aggregate while achieving large gains on reasoning tasks.
\end{itemize}

\section{Related Work}
\label{sec:related_work}

\paragraph{From Uniform to Mixed-Precision Quantization.}
Post-training quantization~(PTQ) compresses LLM weights into low-bit formats without retraining. Standard PTQ methods, whether activation-aware~\citep{lin2024awq,xiao2023smoothquant}, second-order~\citep{frantar2023optq,kim2025guidedquant,zhao2026admmq}, outlier-aware~\citep{dettmers2023spqr,li2026osaq}, or calibration-free~\citep{badri2023hqq}, apply \emph{uniform} precision across layers. More aggressive approaches push into sub-3-bit regimes via randomized transforms~\citep{tseng2024quip}, multi-codebook vector quantization~\citep{egiazarian2024extreme}, or smoothness-aware optimization~\citep{xu2026smoothness}, but fail to exploit inter-layer redundancy. Mixed-precision quantization~(MPQ) improves the compression--quality tradeoff by assigning heterogeneous bit-widths based on layer sensitivity~\citep{dong2019hawq,huang2024slim,zhao2025coopq,deng2026gemq,dumitru2024layerwise} or channel-wise error minimization~\citep{bai2025skim}. However, existing MPQ frameworks rely on perplexity, Hessian trace, or layer-wise reconstruction error as proxies for sensitivity~(see \Cref{tab:related_comparison} in the appendix). We show that these generic proxies systematically misidentify reasoning-critical layers, motivating a multi-objective approach that incorporates downstream task signals.

\paragraph{Task-Aware Calibration and Reasoning.}
Recent studies reveal that complex reasoning degrades disproportionately under quantization~\citep{li2025quantmeetreason,liu2025quanthurts,chang2025inputs}, with \citet{zhou2026failuremodes} further identifying two distinct failure modes, signal degradation and computation collapse, that explain why reasoning tasks are particularly sensitive to aggressive bit reduction. While task-aware allocation methods like TACQ~\citep{xiao2025tacq} attempt to mitigate this via gradient-based circuit discovery, they require expensive backward-pass computations and produce hardware-unfriendly weight-level mixed precision. Other concurrent efforts assign per-task bit-widths using representation-level signals~\citep{levi2025taq,fan2024deeper} or saliency-aware calibration regularization~\citep{wang2026sarqc}, but without addressing calibration data composition. Meanwhile, the impact of calibration data is beginning to be recognized. Standard pipelines adopt WikiText-2 as the default without re-examination~\citep{frantar2023optq}, yet recent observations in pruning~\citep{ji2025beware}, concurrent quantization frameworks such as COLA~\citep{he2025cola}, and multilingual calibration studies~\citep{chimoto2026calibrating} highlight the necessity of domain curation to preserve general capability. Note that \citet{chimoto2026calibrating} focuses on \emph{language} diversity for multilingual models, whereas our work addresses \emph{task-alignment} diversity for reasoning preservation and provides an automatic optimization framework. Concurrently, \citet{okoshi2026towards} independently observe that reasoning and commonsense knowledge exhibit different quantization sensitivity in a QAT setting, validating our motivation from a complementary paradigm.
However, their approach requires expensive fine-tuning with teacher-guided reward losses, does not observe the non-monotonic diversity tradeoff, and applies uniform precision across layers. While these concurrent studies have independently identified aspects of calibration sensitivity, \method{} contributes a unified framework that mathematically formalizes the Alignment-Diversity Tradeoff and automatically searches its optimum via gradient trace alignment, jointly addressing calibration composition and bit allocation with only forward passes and no gradient computation.

\section{Rethinking Calibration in LLM Quantization}
\label{sec:rethink}

Before presenting our method, we challenge two embedded assumptions in standard quantization pipelines.
In \Cref{subsec:sensitivity_gap}, we show that perplexity-based sensitivity misidentifies reasoning-critical layers.
In \Cref{subsec:alignment_diversity}, we reveal a surprising tradeoff in calibration data composition.

\subsection{The Perplexity Illusion}
\label{subsec:sensitivity_gap}

The standard mixed-precision pipeline assigns bit-widths based on a layer's perplexity degradation under quantization, assuming PPL faithfully reflects all downstream capabilities.
To test this assumption, we conduct independent per-layer W3 sensitivity profiling for three distinct objectives, namely PPL, Math~(GSM8K), and ARC~(ARC-Challenge), on LLaMA-3-8B and Qwen2.5-7B~(protocol in \Cref{app:profiling_setup}).

\paragraph{Sensitivity profiles diverge.}
The three sensitivity dimensions exhibit markedly different spatial patterns.
PPL-sensitive layers concentrate heavily at input embeddings and output projections, whereas reasoning-sensitive layers are broadly distributed across middle layers that perform compositional inference~(full heatmaps in \Cref{app:sensitivity_heatmaps}).
This spatial mismatch implies that a PPL-guided allocator would over-protect boundary layers while leaving the reasoning backbone under-protected.

\begin{wraptable}{r}{0.55\textwidth}
\centering
\small
\caption{Kendall rank correlation $\tau$ between per-layer sensitivity vectors under W3 perturbation. PPL sensitivity is essentially uncorrelated with reasoning sensitivity, while Math and ARC sensitivity share a moderate positive correlation on Qwen but not on LLaMA.}
\label{tab:kendall_tau}
\begin{tabular}{l|ccc|ccc}
\toprule
& \multicolumn{3}{c|}{\textbf{LLaMA-3-8B}} & \multicolumn{3}{c}{\textbf{Qwen2.5-7B}} \\
& PPL & Math & ARC & PPL & Math & ARC \\
\midrule
PPL & 1.00 & $-$0.08 & $-$0.20 & 1.00 & 0.02 & 0.38 \\
Math & & 1.00 & 0.20 & & 1.00 & 0.08 \\
ARC & & & 1.00 & & & 1.00 \\
\bottomrule
\end{tabular}
\end{wraptable}

\paragraph{Quantitative divergence.}
\Cref{tab:kendall_tau} reports the Kendall rank correlation $\tau$ between each pair of sensitivity rankings.
On LLaMA-3-8B, the correlation between PPL and Math sensitivity is $\tau = -0.08$~($p\!=\!0.53$), indicating no statistically significant correlation.
The PPL-sensitive ranking is essentially random with respect to reasoning sensitivity.
A set-theoretic analysis further corroborates this finding: the overlap between the top-8 PPL-sensitive and top-8 math-sensitive layers is merely 2 out of 8, matching the random expected overlap of $K^2/L = 2.0$~(\Cref{app:overlap_analysis}).
On Qwen2.5-7B, PPL and Math sensitivity show a similarly negligible correlation of $\tau = 0.02$, while Math and ARC share only weak agreement~($\tau = 0.08$), suggesting model-dependent sensitivity even within the reasoning category.

We term this the \textbf{Perplexity Illusion}: any allocation strategy relying solely on PPL will systematically under-protect the layers most critical for complex reasoning.
This finding is significant because all existing mixed-precision quantization frameworks use PPL or Hessian-based proxies for allocation (\Cref{tab:related_comparison}).

\subsection{The Alignment-Diversity Tradeoff}
\label{subsec:alignment_diversity}

Given that PPL is an inadequate proxy, a natural remedy is to adopt task-specific calibration data, e.g., mathematical text for math reasoning.
Intuitively, this should maximize activation alignment with the target task.
We sweep the calibration mix from pure general data~(WikiText) to pure task data~(GSM8K) and evaluate the actual quantized performance.

\begin{figure}[t]
\centering
\begin{subfigure}[t]{0.45\columnwidth}
    \centering
    \includegraphics[width=1.0\textwidth]{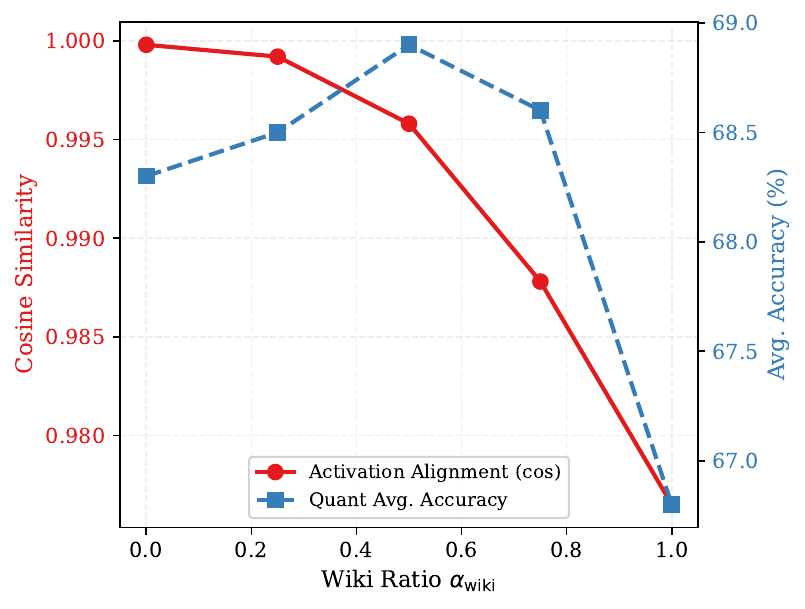}
    \caption{LLaMA-3-8B (peaks around $\alpha_{\text{wiki}}\!\approx\!0.50$).}
    \label{fig:alignment_diversity_llama}
\end{subfigure}
\hspace{1em}
\begin{subfigure}[t]{0.45\columnwidth}
    \centering
    \includegraphics[width=1.0\textwidth]{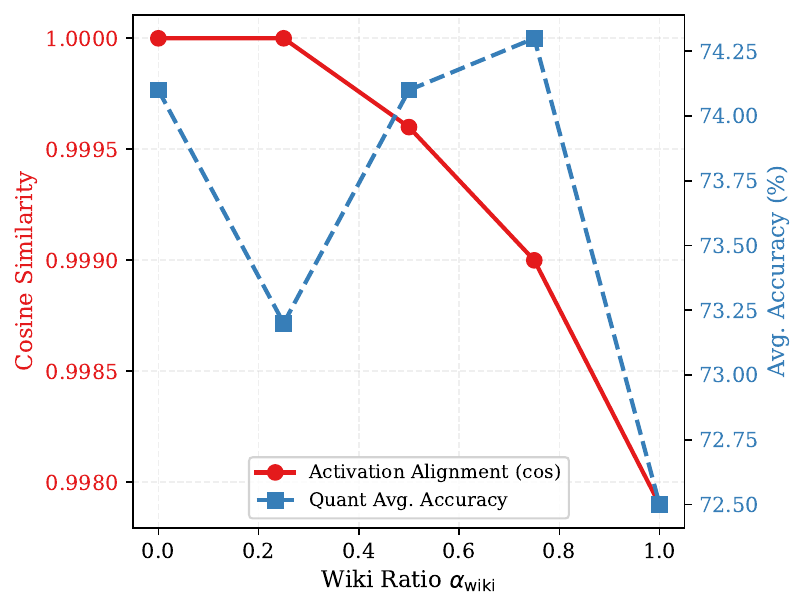}
    \caption{Qwen2.5-7B (peaks around $\alpha_{\text{wiki}}\!\approx\!0.75$).}
    \label{fig:alignment_diversity_qwen}
\end{subfigure}
\caption{\textbf{The Alignment-Diversity Tradeoff.} Red curve~(left axis): cosine similarity between calibration and target-task traces. Blue dashed curve~(right axis): post-quantization aggregate accuracy~(Avg.). Optimal accuracy occurs at an intermediate mixing ratio, not at maximum alignment.}
\label{fig:alignment_diversity}
\end{figure}

\paragraph{Alignment does not imply quantization quality.}
\Cref{fig:alignment_diversity} shows that the activation alignment~(cosine similarity to the target task trace) monotonically increases as we add more task data.
However, the actual quantization quality peaks at an intermediate mixing ratio~($\alpha_{\text{wiki}}\!\approx\!0.50$ on LLaMA-3) and then declines when moving toward pure task data.
We call this divergence the \textbf{Diversity Gap}~(formally defined in \Cref{app:diversity_gap_def}).
The gap is especially pronounced on LLaMA-3, where pure task calibration~($\alpha_{\text{wiki}}\!=\!0$) achieves Avg.$\!=\!$68.3, a full 0.6 points below the balanced mix~(68.9), despite achieving nearly perfect trace alignment~($\cos\!=\!0.9998$).

\paragraph{Diversity as Hessian regularization.}
When calibration data is dominated by a single task distribution, the layer-wise Hessian $\mathbf{H}_l = \mathbf{X}_l^\top \mathbf{X}_l$ develops a degenerate eigenspectrum: a few task-relevant eigenvalues grow large while the rest are suppressed.
Quantization under such a Hessian preserves the narrow task-relevant subspace but distorts the broader representational manifold.
General-domain data lifts the suppressed trailing eigenvalues, maintaining a well-conditioned Hessian that preserves generalization structure.
Empirical eigenspectral analysis confirms this: at reasoning-critical layers, pure math calibration increases the condition number by over $2\times$ and collapses the effective rank from 6 to 3~(\Cref{app:eigenspectral_theory}).

The optimal mixing ratio is both model-specific and task-specific.
We formalize this as \Cref{thm:optimal_mix} in \Cref{app:diversity_gap_def}: under mild assumptions on the Hessian spectra, the expected quantization loss $f(\alpha) = \operatorname{tr}(\mathbf{H}_{\text{test}}\,\mathbf{H}_\alpha^{-1})$ is strictly convex with a \emph{unique} minimum at $\alpha^* \in (0,1)$, ruling out fixed heuristic recipes and motivating our automatic search~(\Cref{subsec:auto_calib}).

\section{Methodology}
\label{sec:method}

\begin{figure}[t]
\centering
\includegraphics[width=0.9\columnwidth]{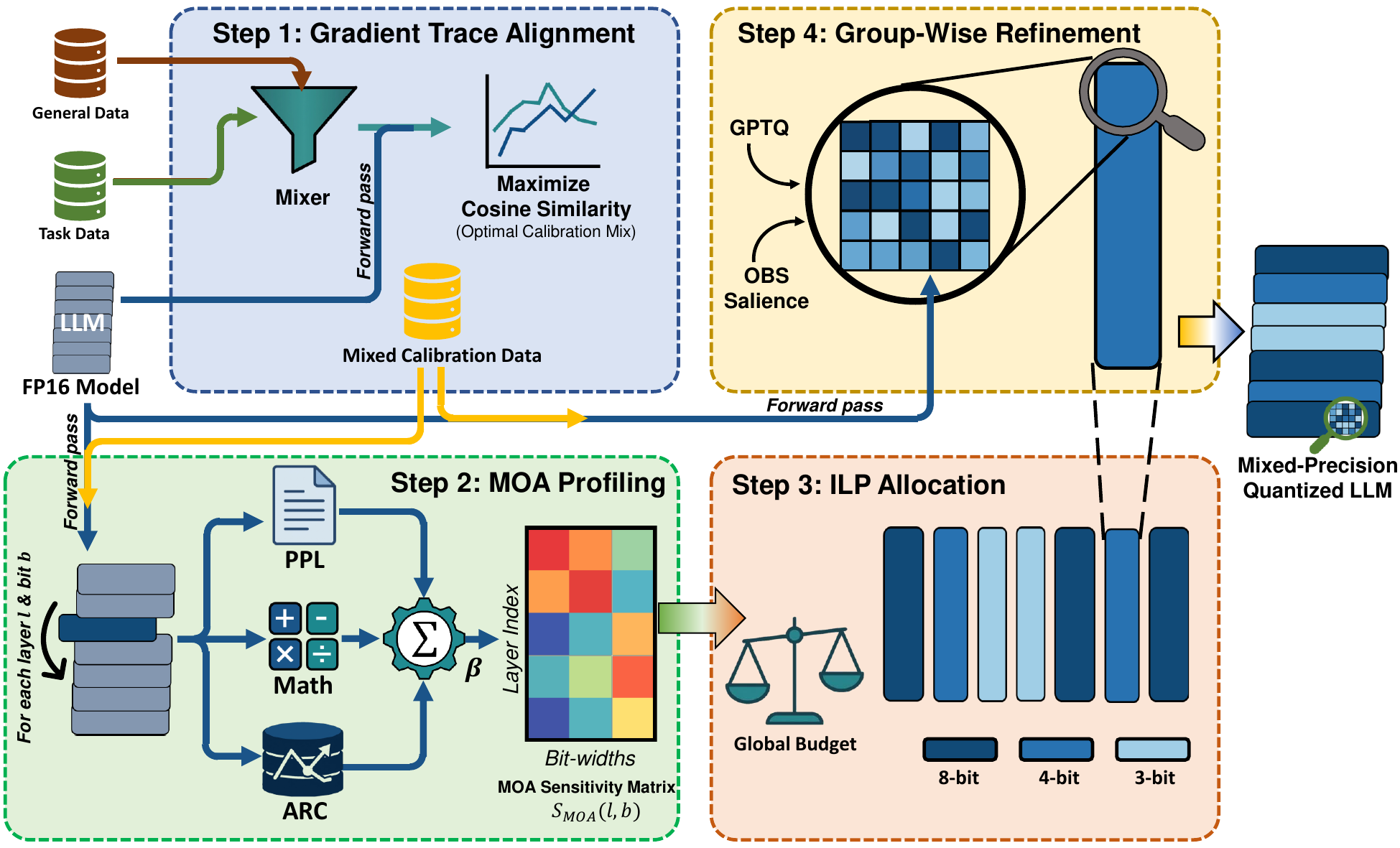}
\caption{\textbf{Overview of the \method{} framework.} \method{} jointly optimizes data composition and bit allocation through four steps: (1) finding the optimal calibration mix via training-free gradient trace alignment; (2) computing a Multi-Objective Aggregation (MOA) sensitivity matrix across task-specific metrics; (3) determining inter-layer average bit-widths via Integer Linear Programming (ILP) under a global budget; and (4) refining intra-layer precision using group-wise OBS salience to yield the final quantized model.}
\label{fig:framework}
\end{figure}

The findings of \Cref{sec:rethink} point to two coupled problems in the standard quantization pipeline: the calibration data is misaligned with the target task, and the sensitivity metric is blind to task-specific objectives.
\method{} addresses both via a two-level framework (\Cref{fig:framework}).
We first describe the automatic calibration strategy~(\Cref{subsec:auto_calib}), then present the inter-layer and intra-layer bit allocation~(\Cref{subsec:moa_alloc,subsec:intra_layer}).

\subsection{Gradient Trace Alignment Auto-Calibration}
\label{subsec:auto_calib}

As established in \Cref{subsec:alignment_diversity}, the optimal calibration mixture is a model- and task-specific intermediate between pure task data and pure generic data.
A brute-force grid search is prohibitively expensive; we instead develop a lightweight proxy that predicts calibration quality from a single forward pass without any actual quantization.

\paragraph{From loss landscape to activation energy.}
The key insight comes from the second-order approximation of quantization loss.
When layer $l$ is quantized from weight $\mathbf{W}_l$ to $\hat{\mathbf{W}}_l = \mathbf{W}_l + \boldsymbol{\delta}_l$, the resulting output degradation is dominated by the quadratic term $\tfrac{1}{2}\,\boldsymbol{\delta}_l^\top \mathbf{H}_l\,\boldsymbol{\delta}_l$, where $\mathbf{H}_l = \mathbf{X}_l^\top \mathbf{X}_l \in \mathbb{R}^{d \times d}$ is the Hessian approximation constructed from the input activation $\mathbf{X}_l \in \mathbb{R}^{n \times d}$ at layer $l$ over $n$ calibration samples.
Rather than computing the full $d \times d$ Hessian, which would be prohibitively expensive for $d\!=\!4096$, we observe that its trace provides a compact energy summary:
\begin{equation}
    \tr(\mathbf{H}_l) = \tr(\mathbf{X}_l^\top \mathbf{X}_l) = \|\mathbf{X}_l\|_F^2,
    \label{eq:trace_frobenius}
\end{equation}
which is simply the squared Frobenius norm of the activation matrix.
Layers with larger trace values have larger-magnitude activations and are thus more susceptible to quantization error, since the quadratic penalty scales with the eigenvalues of $\mathbf{H}_l$, whose sum equals the trace.

We collect these per-layer traces into a \emph{trace vector} that summarizes the activation energy landscape of the entire model under a given data distribution:
\begin{equation}
    \mathbf{h}(\mathcal{D}) = \left(\tr(\mathbf{X}_1^\top \mathbf{X}_1), \;\ldots,\; \tr(\mathbf{X}_L^\top \mathbf{X}_L)\right) \in \mathbb{R}^L.
    \label{eq:trace_vector}
\end{equation}

\paragraph{Optimization objective.}
Given a mixed distribution $\mathcal{D}_{\text{mix}}(\alpha)$ blending general data~(WikiText) and task data at ratio $\alpha$ ($\alpha\!=\!1$: pure WikiText; $\alpha\!=\!0$: pure task), we select the ratio maximizing trace alignment with the target task:
\begin{equation}
    \alpha^* = \argmax_{\alpha \in [0,1]} \;\cos\!\big(\mathbf{h}(\mathcal{D}_{\text{mix}}(\alpha)),\; \mathbf{h}(\mathcal{D}_{\text{task}})\big).
    \label{eq:auto_calib}
\end{equation}
Cosine similarity is preferred over Euclidean distance for its scale invariance: the \emph{relative} distribution of energy across layers matters more than the absolute magnitude.
Solving Eq.~\eqref{eq:auto_calib} requires only $(|\mathcal{A}|+1)$ forward passes with $n\!=\!16$ samples per candidate.
Note that trace alignment operates at the macroscopic inter-layer level: it routes activation energy across layers without constraining the intra-layer spectral structure that diversity protects~(\Cref{app:auto_calib_details}), so maximizing Eq.~\eqref{eq:auto_calib} does not conflict with the tradeoff discussed in \Cref{subsec:alignment_diversity}.
In practice, the cosine landscape is smooth, so a coarse grid $\alpha \in \{0.0, 0.25, 0.5, 0.75, 1.0\}$ suffices~(\Cref{app:auto_calib_details}).

\subsection{Level 1: Inter-Layer Allocation via MOA}
\label{subsec:moa_alloc}

With the calibration composition determined, we turn to the bit allocation problem.

\paragraph{Multi-Objective Aggregation (MOA).}
For each layer $l$ and candidate bit-width $b \in \mathcal{B} = \{3, 4, 8\}$, we measure the sensitivity as the relative degradation in multiple independent metrics when layer $l$ alone is quantized to $b$ bits while all other layers remain at FP16:
\begin{equation}
    \Delta_l^{(k)}(b) = \frac{M_k(\theta_{l \to b}) - M_k(\theta)}{M_k(\theta)}, \quad k \in \{\text{ppl}, \text{math}, \text{arc}\},
    \label{eq:per_layer_sens}
\end{equation}
where $\theta_{l \to b}$ denotes the model with layer $l$ quantized to $b$ bits and $M_k$ is the evaluation metric for objective $k$.
All three metrics are loss-type~(higher is worse): $M_{\text{ppl}}$ is perplexity, while $M_{\text{math}}$ and $M_{\text{arc}}$ are conditional cross-entropy losses on GSM8K and ARC-Challenge respectively.
Using cross-entropy rather than accuracy avoids autoregressive generation and reduces each profiling pass to a single forward evaluation; this follows the standard OBS assumption that loss-based sensitivity faithfully tracks downstream degradation~\citep{frantar2023optq,hassibi1993obs}.
Thus $\Delta_l^{(k)} > 0$ uniformly indicates degradation under quantization.
The ARC sensitivity is used in our analysis of the Perplexity Illusion~(\Cref{subsec:sensitivity_gap}) but is not included in the allocation objective, since Math and ARC sensitivity share substantial overlap in their layer rankings~(\Cref{app:overlap_analysis}).
This profiling requires $L \times |\mathcal{B}|$ forward passes, one per layer per bit-width candidate.

The composite MOA score aggregates the PPL and Math signals as a weighted sum:
\begin{equation}
    S_{\text{MOA}}(l, b) = \beta \cdot \Delta_l^{\text{math}}(b) + (1\!-\!\beta) \cdot \Delta_l^{\text{ppl}}(b),
    \label{eq:moa}
\end{equation}
where $\beta$ controls the balance between task-specific protection and general language modeling preservation.
We set $\beta\!=\!0.7$ by default, giving dominant weight to the task-specific signal, which is the primary motivation of \method{}.
As validated in \Cref{app:beta_ablation}, Avg.\ varies by only 0.5 percentage points across $\beta \in [0.5, 0.9]$, suggesting that the main contribution of MOA lies in the structural inclusion of task-aware signals rather than the precise tuning of their weights.

\paragraph{Integer Linear Programming (ILP).}
We minimize the aggregate MOA sensitivity subject to a global bit-budget constraint $B$:
\begin{equation}
    \min_{\{b_l\}_{l=1}^L} \sum_{l=1}^{L} S_{\text{MOA}}(l, b_l), \quad \text{s.t.} \quad \frac{1}{L}\sum_{l=1}^{L} b_l \leq B, \quad b_l \in \mathcal{B}.
    \label{eq:ilp}
\end{equation}
We exploit the problem's sequential structure to solve this ILP exactly via dynamic programming in $O(L \cdot B_{\text{total}} \cdot |\mathcal{B}|)$ time, which is negligible ($<\!1$ second for all configurations; recurrence details in \Cref{app:dp_details}).
The resulting allocation produces a heterogeneous per-layer bit pattern that interleaves 3-bit, 4-bit, and occasionally 8-bit layers according to the multi-objective sensitivity landscape.
The W4 layers concentrate at the union of PPL-sensitive and reasoning-sensitive positions, simultaneously protecting both types of critical layers~(\Cref{tab:app_allocation}).

\subsection{Level 2: Intra-Layer Group-Wise Allocation}
\label{subsec:intra_layer}

While the inter-layer ILP determines each layer's \emph{average} bit-width, substantial sensitivity variation exists \emph{within} each layer across different weight groups.
To exploit this residual heterogeneity, we introduce a second level of mixed-precision allocation at the group level.
Following SliM-LLM~\citep{huang2024slim}, we apply GPTQ~\citep{frantar2023optq} with OBS-based salience scoring to rank weight groups~($g\!=\!128$) by their downstream impact.
Each group is then assigned one of three precision levels~($b_l\!-\!1$, $b_l$, or $b_l\!+\!1$ bits), subject to the constraint that the group-level average matches the layer budget $b_l$.
This two-level architecture enables compression rates impossible with layer-level allocation alone: a 4-bit layer may contain 3-bit and 5-bit groups, concentrating precision on the most critical subsets~(\Cref{app:intra_layer_details}).

\subsection{Computational Cost}
\label{subsec:cost}

A core design choice in \method{} is the strict decoupling of the task-aware search overhead from the backend quantization execution.
The \method{}-specific offline pipeline takes approximately \textbf{47 minutes} on a single A100 for LLaMA-3-8B~(\Cref{app:timing}), comparable to existing mixed-precision methods.
This is a one-time cost; new bit budgets require only re-solving the ILP~($<\!1$s).

\section{Experiments}
\label{sec:experiments}

We evaluate \method{} across two model families, eight benchmarks, and multiple bit budgets.
Our experiments address four questions:
(1)~Does \method{} outperform state-of-the-art baselines?
(2)~How much does allocation strategy matter versus calibration data composition?
(3)~What is the optimal calibration mix?
(4)~How does performance scale with bit budget?

\paragraph{Setup.}
We evaluate on LLaMA-3-8B~\citep{grattafiori2024llama}~(32 layers) and Qwen2.5-7B~\citep{yang2024qwen2}~(28 layers).
Benchmarks span four categories: reasoning~(GSM8K 8-shot CoT, ARC-Challenge 25-shot, ARC-Easy), commonsense~(HellaSwag 10-shot, WinoGrande 5-shot, PIQA), reading comprehension~(BoolQ), and language modeling~(WikiText-2 PPL).
We report Avg., the unweighted mean of the seven accuracy-based tasks, as the primary aggregate metric.
All evaluations use lm-evaluation-harness~\citep{eval-harness} on full test sets.
We compare against seven baselines spanning four paradigms: activation-aware~(AWQ), second-order~(GPTQ, SpQR), calibration-free~(HQQ, OWQ), and naive~(RTN), all with $g\!=\!128$ and WikiText-2 calibration~(128 samples).
\method{} uses MOA with $\beta\!=\!0.7$, mixed calibration~($\alpha^*_{\text{wiki}}$ from auto-calibration with a multi-task reference trace combining GSM8K and ARC-Challenge; defaults: $\alpha_{\text{wiki}}\!=\!0.50$ for LLaMA-3, $\alpha_{\text{wiki}}\!=\!0.75$ for Qwen2.5), $g\!=\!128$, $n\!=\!128$, and seqlen$\!=\!$2048.
Full details in \Cref{app:impl}.

\begin{table*}[t]
\centering
\caption{Allocation strategy ablation~(LLaMA-3-8B, b3.5). \textbf{Bold}: best in column; \underline{underline}: second best. The allocation strategy provides the dominant improvement~(+3.5 Avg.\ over Random, +5.6 over PPL-only), while mixed calibration provides an additive boost~(+2.1 Avg.\ for MOA, +5.1 for Uniform). $^\dagger$Uniform+wiki is equivalent to SliM-LLM~\citep{huang2024slim} with uniform inter-layer precision, serving as the mixed-precision baseline.}
\label{tab:strategy_ablation}
\small
\setlength{\tabcolsep}{1pt}
\begin{tabular}{llccccccccc}
\toprule
\textbf{Strategy} & \textbf{Calib} & \textbf{ARC-C} & \textbf{HSwag} & \textbf{WinoG} & \textbf{PIQA} & \textbf{ARC-E} & \textbf{BoolQ} & \textbf{GSM8K} & \textbf{PPL}$\downarrow$ & \textbf{Avg.}$\uparrow$ \\
\midrule
MOA & mixed & \underline{50.0} & \textbf{76.9} & \textbf{73.6} & 78.6 & 77.7 & \underline{79.6} & \textbf{46.2} & 8.72 & \textbf{68.9} \\
MOA & wiki & 48.5 & \underline{76.7} & 72.8 & \textbf{78.8} & \textbf{78.7} & 78.1 & 33.7 & \textbf{8.59} & 66.8 \\
MOA-norm & mixed & \textbf{50.8} & \underline{76.7} & \underline{73.1} & \underline{78.7} & \underline{78.3} & 79.4 & \underline{44.4} & \underline{8.71} & \underline{68.8} \\
KBA-Math & mixed & 48.7 & 76.1 & 73.0 & 78.5 & 77.3 & \textbf{79.7} & 43.4 & 8.99 & 68.1 \\
PPL-topK & wiki & 44.9 & 74.6 & \underline{73.1} & 77.3 & 73.9 & 76.2 & 23.1 & 9.14 & 63.3 \\
Uniform & mixed & 48.5 & 75.2 & 72.5 & 78.2 & 77.0 & 77.2 & 39.5 & 9.29 & 66.9 \\
Uniform$^\dagger$ & wiki & 44.2 & 75.7 & 70.0 & 75.4 & 73.6 & 70.1 & 23.3 & 9.18 & 61.8 \\
Random~(3 seeds) & wiki & $47.4 {\scriptstyle \pm 2.3}$ & $75.9 {\scriptstyle \pm 0.5}$ & $73.0 {\scriptstyle \pm 0.2}$ & $76.8 {\scriptstyle \pm 1.0}$ & $75.8 {\scriptstyle \pm 0.7}$ & $78.1 {\scriptstyle \pm 1.3}$ & $30.6 {\scriptstyle \pm 1.2}$ & $8.97 {\scriptstyle \pm 0.28}$ & 65.4 \\
\bottomrule
\end{tabular}
\end{table*}

\begin{figure}[t]
\centering
\begin{subfigure}[t]{0.45\columnwidth}
    \centering
    \includegraphics[width=1.0\textwidth]{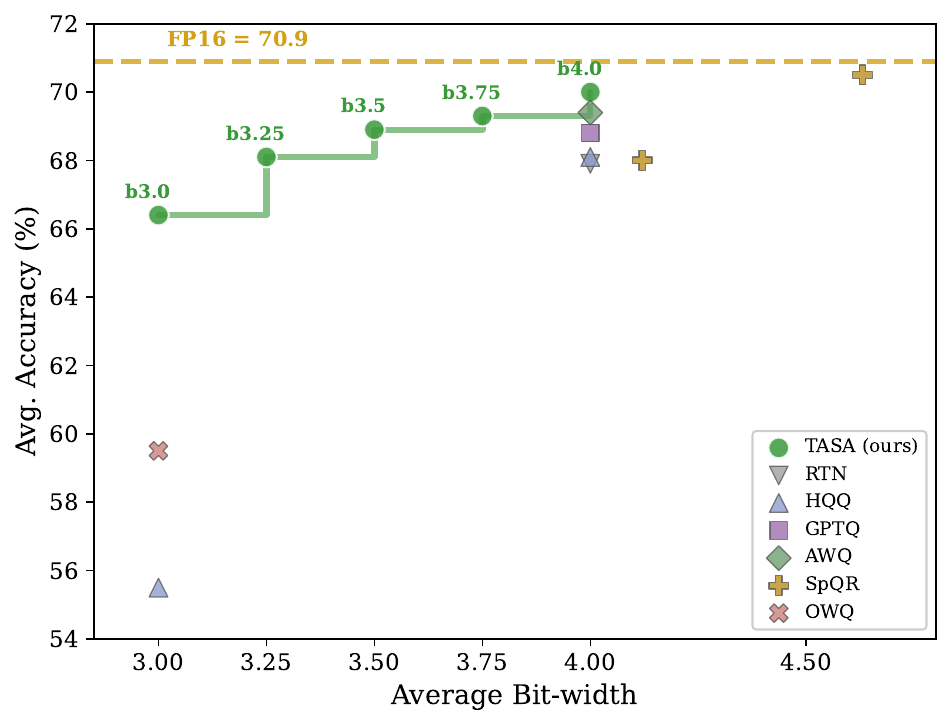}
    \caption{LLaMA-3-8B}
    \label{fig:main_results_llama}
\end{subfigure}
\hspace{1em}
\begin{subfigure}[t]{0.45\columnwidth}
    \centering
    \includegraphics[width=1.0\textwidth]{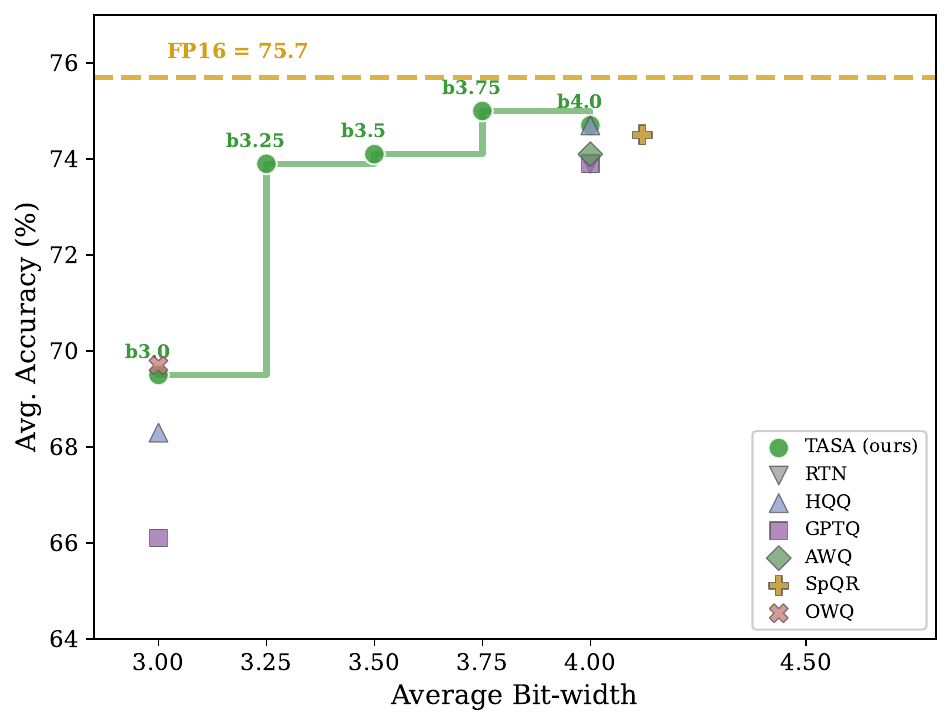}
    \caption{Qwen2.5-7B}
    \label{fig:main_results_qwen}
\end{subfigure}
\caption{\textbf{Aggregate accuracy vs.\ effective bit-width.} The green scaling curve illustrates \method{}'s smooth performance progression from 3.0 to 4.0 bits. Gray markers denote uniform-precision baselines at their effective bit-widths, and the dashed line represents the FP16 upper bound. 
}
\label{fig:main_results}
\end{figure}

\subsection{Main Results}
\label{subsec:main_results}

\Cref{tab:main_llama,tab:main_qwen} in \Cref{app:full_results} present the full per-task results; \Cref{fig:main_results} provides a visual summary.
\method{} achieves \emph{precision inversion}: fewer bits yield better reasoning performance than uniform baselines at higher precision.

\paragraph{Precision inversion at 3.5 bits.}
\method{} b3.5 on LLaMA-3~(Avg.$\!=\!$68.9) matches 4-bit uniform baselines, including HQQ W4~(68.1), RTN W4~(67.9), and even approaches AWQ W4~(69.4), while using 12.5\% fewer bits on average.
The advantage is especially pronounced on GSM8K, where \method{} b3.5 achieves 46.2, compared to HQQ W4's 39.4~(+17\%) and RTN W4's 36.2~(+28\%).
On Qwen2.5, \method{} b3.75~(Avg.$\!=\!$75.0) retains 99.1\% of FP16 accuracy~(75.7), exceeding the best 4-bit uniform methods including HQQ W4~(74.7) and AWQ W4~(74.1) while using 6.25\% fewer bits.
We also observe that \method{} b3.75 achieves the highest ARC-C~(53.9 vs.\ FP16's 51.1) among all configurations, suggesting that heterogeneous precision allocation may act as an implicit regularizer on classification-style tasks.

\paragraph{Dominance at W3.}
At 3 bits, the gap widens substantially.
On LLaMA-3, \method{} b3.0~(Avg.$\!=\!$66.4, GSM8K$\!=\!$39.5) outperforms HQQ W3~(55.5, 10.0) by +10.9 Avg.\ and +29.5 GSM8K, OWQ W3~(59.5, 19.2) by +6.9 and +20.3, and RTN W3~(47.3, 0.0) by +19.1 Avg.
The near-zero GSM8K scores for HQQ and RTN at W3 indicate severe reasoning collapse, a failure mode that \method{}'s task-aware allocation avoids.
SpQR ``W3'' achieves Avg.$\!=\!$68.0, but its effective bit-width is 4.12 due to metadata and FP16 outlier overhead~(\Cref{app:spqr_bits}).

\paragraph{Per-task analysis.}
Across both models, \method{} exhibits a consistent pattern.
The largest improvements relative to uniform baselines occur on reasoning tasks~(GSM8K, ARC-C), moderate improvements on commonsense tasks~(HellaSwag, WinoGrande, PIQA), and competitive but not dominant performance on BoolQ.
MOA allocates higher precision to reasoning-sensitive layers.
Comparison with task-specific methods such as TACQ~\citep{xiao2025tacq} is provided in \Cref{app:tacq}.

\subsection{Allocation Strategy Ablation}
\label{subsec:strategy_ablation}

A central question is whether \method{}'s advantage stems primarily from the allocation strategy~(MOA versus alternatives) or from the calibration data composition~(mixed versus pure WikiText).
To disentangle them, we conduct a factorial ablation on LLaMA-3-8B at b3.5, as shown in \Cref{tab:strategy_ablation}.

\emph{\textbf{(i)}~MOA dominates PPL-only allocation.}
Comparing MOA+mixed~(Avg.$\!=\!$68.9) with PPL-topK+wiki~(63.3) reveals a +5.6 point improvement.
On GSM8K, the gap reaches +23.1~(46.2 vs.\ 23.1), showing that PPL-guided allocation systematically under-protects reasoning-critical layers.
This is the largest single-factor improvement in the ablation suite.

\emph{\textbf{(ii)}~Mixed calibration provides an additive boost.}
Holding the allocation strategy fixed, switching from wiki-only to mixed calibration improves performance across all strategies.
MOA gains +2.1 Avg.\ and Uniform gains +5.1 Avg.
On GSM8K, the improvement from mixed calibration is especially large: +12.5 points for MOA~(46.2 vs.\ 33.7) and +16.2 points for Uniform~(39.5 vs.\ 23.3), suggesting that task-aligned calibration data partially compensates for the lack of task-aware allocation.
The larger calibration gain under Uniform allocation (+5.1) than under MOA (+2.1) reflects a substitution effect: when the allocation strategy already protects reasoning-critical layers, the marginal value of improved calibration data diminishes; conversely, when allocation is task-agnostic, calibration quality becomes the primary lever for preserving reasoning.

\emph{\textbf{(iii)}~Random allocation collapses on reasoning.}
Random allocation with WikiText calibration achieves Avg.$\!=\!$65.4, which is competitive on commonsense tasks.
However, its GSM8K score~(30.6) is much lower than MOA~(46.2), revealing that random allocation cannot protect the layers critical for multi-step reasoning.
KBA-Math~(single-objective, Avg.$\!=\!$68.1) is strong, but MOA provides an additional +0.8 by jointly optimizing for language modeling and reasoning.

\subsection{Calibration Data Ablation}
\label{subsec:calib_ablation}

\begin{table}[t]
\centering
\caption{Mix ratio ablation~(b$\!=\!$3.5, MOA). Performance peaks at intermediate mixing ratios for both models, confirming the Alignment-Diversity Tradeoff. Neither pure task data~($\alpha_{\text{wiki}}\!=\!0$) nor pure generic data~($\alpha_{\text{wiki}}\!=\!1$) is optimal. 
}
\label{tab:mix_ratio}
\resizebox{\columnwidth}{!}{
\begin{tabular}{lcccccccccc}
\toprule
\textbf{Model} & $\alpha_{\text{wiki}}$ & \textbf{ARC-C} & \textbf{HSwag} & \textbf{WinoG} & \textbf{PIQA} & \textbf{ARC-E} & \textbf{BoolQ} & \textbf{GSM8K} & \textbf{PPL}$\downarrow$ & \textbf{Avg.}$\uparrow$ \\
\midrule
\multirow{5}{*}{LLaMA-3} & 0.00 & 49.5 & 76.0 & 73.6 & 78.3 & 77.0 & 79.0 & 44.6 & 9.41 & 68.3 \\
& 0.25 & 49.8 & \textbf{77.0} & 73.2 & 78.2 & 77.7 & 78.7 & 44.9 & 8.82 & 68.5 \\
& 0.50 & 50.0 & 76.9 & \textbf{73.6} & \textbf{78.6} & 77.7 & \textbf{79.6} & \textbf{46.2} & 8.72 & \textbf{68.9} \\
& 0.75 & \textbf{51.6} & 76.5 & 73.3 & 78.9 & \textbf{78.2} & 79.1 & 42.2 & 8.60 & 68.6 \\
& 1.00 & 48.5 & 76.7 & 72.8 & 78.8 & 78.7 & 78.1 & 33.7 & \textbf{8.59} & 66.8 \\
\midrule
\multirow{5}{*}{Qwen} & 0.00 & 50.6 & 76.7 & 70.4 & 77.8 & 79.6 & 82.9 & \textbf{80.6} & 10.01 & 74.1 \\
& 0.25 & 49.2 & \textbf{77.1} & 69.3 & 77.6 & 78.1 & 82.0 & 78.9 & 9.56 & 73.2 \\
& 0.50 & 50.6 & 77.0 & \textbf{70.6} & 77.9 & 79.1 & \textbf{84.6} & 79.2 & 9.47 & 74.1 \\
& 0.75 & \textbf{52.1} & 76.5 & 70.2 & \textbf{77.9} & \textbf{79.9} & 83.4 & 79.9 & 9.44 & \textbf{74.3} \\
& 1.00 & 50.9 & 76.8 & 68.4 & 78.0 & 78.6 & 83.1 & 71.9 & \textbf{9.42} & 72.5 \\
\bottomrule
\end{tabular}
}
\end{table}

\Cref{tab:mix_ratio} presents the mix ratio ablation at b3.5 with MOA allocation on both models, confirming that the optimal mix ratio is model-specific.
On LLaMA-3, intermediate ratios~($\alpha_{\text{wiki}} \in [0.25, 0.50]$) form a competitive plateau for GSM8K, with the balanced mix~($\alpha_{\text{wiki}}\!=\!0.50$) achieving the highest score~(46.2).
On Qwen2.5, $\alpha_{\text{wiki}} \in [0.50, 0.75]$ yields near-equivalent Avg.~(74.1 to 74.3), with $\alpha_{\text{wiki}}\!=\!0.75$ marginally ahead; this model benefits more from calibration diversity, consistent with its math-enriched pre-training.
Neither pure task data~($\alpha_{\text{wiki}}\!=\!0$) nor pure generic data~($\alpha_{\text{wiki}}\!=\!1$) is optimal on either model, providing direct experimental confirmation of the Alignment-Diversity Tradeoff.
The auto-calibration proxy~(\Cref{subsec:auto_calib}) reliably narrows the search to the competitive neighborhood at negligible cost~(full results in \Cref{app:auto_calib_results}; per-task visualizations in \Cref{app:mix_bar}).

\subsection{Budget Sweep}
\label{subsec:budget_sweep}

\Cref{fig:main_results} plots \method{}'s aggregate performance from b3.0 to b4.0, revealing smooth scaling on both models~(\Cref{app:budget_sweep}).
On LLaMA-3, b3.5 captures 97.2\% of FP16 accuracy~(68.9 vs.\ 70.9) at $4.57\times$ compression.
On Qwen2.5, b3.75~(Avg.$\!=\!$75.0) slightly exceeds b4.0~(74.7).
Reasoning tasks degrade more steeply under aggressive compression, but \method{} flattens this curve: on LLaMA-3 at b3.0 it achieves 39.5\% on GSM8K versus 10--19\% for HQQ/OWQ.

\section{Conclusion}
\label{sec:conclusion}

We identified two blind spots in standard LLM quantization: the \emph{Perplexity Illusion}, where PPL-based sensitivity fails to correlate with reasoning capabilities, and the \emph{Alignment-Diversity Tradeoff}, where calibrating exclusively on task-specific data degrades generalization.
To address both, we introduced \method{}, a two-level task-aware quantization framework combining training-free gradient trace alignment for calibration-data search with a Multi-Objective Aggregation (MOA) metric for structured inter- and intra-layer bit allocation.
On LLaMA-3-8B and Qwen2.5-7B, \method{} achieves \emph{precision inversion}, matching several 4-bit uniform baselines at only 3.5-bit average precision while substantially improving reasoning accuracy over uniform W3 methods.

\textbf{Limitations.} Our experiments focus on 7--8B models; scaling MOA profiling to larger architectures would require distributed computation, though per-layer independence makes this easy to parallelize. The auto-calibration searches a discrete grid of mixing ratios; continuous relaxation could yield finer optima. The heterogeneous bit patterns are not yet natively supported by all inference kernels, though frameworks like MLC-LLM and vLLM already handle mixed-precision decoding.

\textbf{Broader Impacts.} By preserving reasoning capabilities under aggressive quantization, \method{} enables efficient LLM deployment on resource-constrained devices, reducing computational costs and energy consumption.

\newpage
\bibliographystyle{iclr2026_conference}
\bibliography{ref}

\newpage
\appendix
\section{Experimental Setup}
\label{app:details}

\subsection{Method Comparison}
\label{app:method_comparison}

\begin{table}[t]
\centering
\caption{Comparison of mixed-precision quantization approaches along three design dimensions. \method{} jointly optimizes the sensitivity metric, calibration data, and allocation granularity in a task-aware manner.}
\label{tab:related_comparison}
\small
\setlength{\tabcolsep}{4pt}
\begin{tabular}{lcccc}
\toprule
\textbf{Method} & \textbf{Sensitivity Metric} & \textbf{Calib Data} & \textbf{Granularity} & \textbf{Task-Aware?} \\
\midrule
HAWQ & Hessian eigenvalue & Generic & Layer & No \\
HAWQ-V2 & Hessian trace & Generic & Layer & No \\
OWQ & Activation outliers & Generic & Column & No \\
SliM-LLM & OBS salience & Generic & Group & No \\
APTQ & Attention entropy & Generic & Layer & No \\
CoopQ & Shapley value & Generic & Layer & No \\
TACQ & Circuit discovery & Task & Weight & Yes \\
TAQ & Hidden repr. & Task & Layer & Yes \\
\midrule
\textbf{\method{}} & \textbf{MOA (multi-obj)} & \textbf{Auto-mixed} & \textbf{Layer+Group} & \textbf{Yes} \\
\bottomrule
\end{tabular}
\end{table}

\Cref{tab:related_comparison} compares \method{} with existing mixed-precision quantization methods along three design dimensions: sensitivity metric, calibration data, and allocation granularity. \method{} jointly optimizes all three dimensions in a task-aware manner.

\subsection{Implementation Details}
\label{app:impl}

All experiments are conducted on NVIDIA A100-SXM4-40GB GPUs with CUDA 12.1 and PyTorch 2.1.
Models are loaded in FP16 precision from local checkpoints, specifically Meta-LLaMA-3-8B~(32 decoder layers, $d_{\text{model}}\!=\!4096$, GQA with 8 KV heads, 32k vocabulary) and Qwen2.5-7B~(28 decoder layers, $d_{\text{model}}\!=\!3584$, GQA with 4 KV heads, 152k vocabulary).

Quantization hyperparameters are fixed across all configurations: group size $g\!=\!128$, calibration set size $n_{\text{samples}}\!=\!128$, sequence length 2048, and random seed 42 for reproducibility.
The GPTQ implementation follows SliM-LLM's codebase with OBS-based group-wise salience for intra-layer allocation.
For the auto-calibration search, we use $n\!=\!16$ samples per candidate ratio, which provides sufficient stability for the Frobenius norm computation.
For \method{}, the generic calibration component draws 128 samples from the WikiText-2 training partition; the task-specific component draws from the GSM8K training set~(7,473 examples).
In the sensitivity profiling stage, PPL is measured on the WikiText-2 validation set, while the conditional cross-entropy losses for Math and ARC are computed on samples from the GSM8K and ARC-Challenge training sets, respectively.
The auto-calibration target trace uses a separate random seed to ensure the reference samples are disjoint from the calibration data.

Evaluation is performed using lm-evaluation-harness~\citep{eval-harness} with the following few-shot settings: GSM8K~(8-shot, chain-of-thought), ARC-Challenge~(25-shot), HellaSwag~(10-shot), WinoGrande~(5-shot), and all other tasks in zero-shot.
All evaluations use the \emph{complete} test sets with no sample limits.

For sensitivity profiling, we quantize each layer independently to each candidate bit-width in $\mathcal{B}\!=\!\{3, 4, 8\}$ using group-wise RTN~($g\!=\!128$), measure the three sensitivity metrics~(perplexity on WikiText-2, conditional cross-entropy loss on GSM8K 0-shot, conditional cross-entropy loss on ARC-Challenge 0-shot), and restore the layer to FP16 before proceeding to the next.
The profiling uses $n\!=\!64$ calibration samples per metric.
Full details are provided in \Cref{app:profiling_setup}.

\subsection{Detailed Sensitivity Profiling Protocol}
\label{app:profiling_setup}

To measure how layer sensitivity diverges across different evaluation objectives, we conduct independent per-layer sensitivity analyses.
Given a model with $L$ layers, we quantize each layer $l$ individually to 3-bit precision using group-wise round-to-nearest~(RTN, group size $g\!=\!128$) while keeping all other layers at FP16, and measure the resulting degradation in three metrics.
We deliberately use RTN rather than GPTQ for profiling: RTN applies a pure rounding perturbation without second-order error compensation, so the measured degradation isolates each layer's intrinsic sensitivity to weight perturbation rather than conflating it with the compensatory capacity of a specific quantizer.
This choice also reduces profiling cost from one Hessian construction per layer to a single rounding pass.
The resulting sensitivity rankings are:
\begin{itemize}
    \item \textbf{PPL sensitivity} $s_l^{\text{ppl}}$: the relative perplexity increase on WikiText-2, defined as $s_l^{\text{ppl}} = (\text{PPL}_{l \to 3} - \text{PPL}_{\text{fp16}}) / \text{PPL}_{\text{fp16}} \times 100\%$.
    \item \textbf{Math sensitivity} $s_l^{\text{math}}$: the relative increase in conditional cross-entropy loss on GSM8K~(0-shot, teacher-forced), defined analogously.
    \item \textbf{ARC sensitivity} $s_l^{\text{arc}}$: the relative increase in conditional cross-entropy loss on ARC-Challenge~(0-shot, teacher-forced).
\end{itemize}

After measuring each layer, we restore it to FP16 before proceeding to the next.
This ensures that the sensitivity of each layer is measured independently, without interference from other quantized layers.
The profiling uses $n\!=\!64$ calibration samples per metric at a sequence length of 2048.
This procedure yields a sensitivity vector $\mathbf{s}^{(k)} \in \mathbb{R}^L$ for each objective $k \in \{\text{ppl}, \text{math}, \text{arc}\}$, where each entry records the contribution of layer $l$ to objective $k$ under quantization perturbation.

\begin{table}[t]
\centering
\caption{Baseline implementation details. All methods use group size $g\!=\!128$ and WikiText-2 calibration with 128 samples.}
\label{tab:app_baseline_impl}
\resizebox{\columnwidth}{!}{
\begin{tabular}{lll}
\toprule
\textbf{Method} & \textbf{Implementation} & \textbf{Notes} \\
\midrule
RTN & Custom (symmetric MinMax) & Standard round-to-nearest with per-group scales \\
GPTQ & Custom (pure PyTorch) & Standard GPTQ algorithm~\citep{frantar2023optq}, asymmetric \\
AWQ & AutoAWQ v0.2.9 & Official package~\citep{lin2024awq} \\
HQQ & HQQ v0.2.8 & Official package~\citep{badri2023hqq} \\
SpQR & Official codebase & \citet{dettmers2023spqr}, group size 16, bilevel 3-bit \\
OWQ & Official codebase & \citet{lee2024owq}, default configuration \\
SliM-LLM & Official codebase & \citet{huang2024slim}, integrated into our pipeline \\
\bottomrule
\end{tabular}
}
\end{table}

\paragraph{Baseline implementations.}
\Cref{tab:app_baseline_impl} lists the implementation source for each baseline method.
All baselines use the same group size~($g\!=\!128$), calibration data~(WikiText-2, $n\!=\!128$ samples), and evaluation pipeline~(lm-evaluation-harness, full test sets) as \method{} to ensure a fair comparison.

RTN and GPTQ are implemented in-house for compatibility with our evaluation framework~(transformers~4.57+), as the original GPTQ codebase targets older transformer versions.
Our GPTQ implementation follows the standard algorithm of \citet{frantar2023optq} with Cholesky-based Hessian inversion, blockwise error compensation~(block size~128), and per-group asymmetric quantization.
SpQR and OWQ are run from their official repositories with default configurations.
AWQ and HQQ use their official Python packages.
HQQ~\citep{badri2023hqq} does not have a peer-reviewed publication. However, it is widely adopted in practice and officially integrated into the HuggingFace Transformers ecosystem, making it a representative calibration-free baseline.
SliM-LLM's group-wise salience-based GPTQ is integrated into our quantization pipeline for the intra-layer allocation stage described in \Cref{subsec:intra_layer}, and is also used as a standalone baseline by setting the inter-layer allocation to uniform.

\subsection{Auto-Calibration Details}
\label{app:auto_calib_details}

\paragraph{Why cosine similarity?}
We choose cosine similarity over Euclidean distance or KL divergence because it is scale-invariant. Two distributions that produce proportional trace vectors with the same shape but different magnitude achieve $\cos\!=\!1$.
This is desirable because the \emph{relative} distribution of activation energy across layers, which determines where quantization error concentrates, matters more than the absolute magnitude, which depends on factors like batch size and sequence length that are irrelevant to calibration quality.

\paragraph{Why trace alignment preserves diversity.}
At first glance, maximizing the alignment between calibration and target-task traces~(Eq.~\eqref{eq:auto_calib}) may seem to conflict with the Alignment-Diversity Tradeoff of \Cref{subsec:alignment_diversity}, which warns against excessive task alignment.
The resolution lies in recognizing that these two phenomena operate at different granularities.
The trace vector $\mathbf{h}(\mathcal{D})$ captures the \emph{macroscopic} distribution of activation energy across layers, summarizing \emph{which layers} carry the most sensitivity to quantization error.
The diversity requirement, by contrast, concerns the \emph{microscopic} structure within each layer's Hessian $\mathbf{H}_l = \mathbf{X}_l^\top \mathbf{X}_l$, specifically whether its eigenspectrum is broad enough to represent the full weight space.
Since the trace equals the sum of eigenvalues, two Hessians can share the same trace while possessing vastly different spectral profiles: one may be near-singular with energy concentrated in a few directions, while the other distributes energy broadly.
Therefore, maximizing trace alignment guides inter-layer energy routing without constraining the intra-layer spectral quality that diversity protects.

\paragraph{Practical considerations.}
The cosine similarity landscape is smooth, so a coarse grid of $\alpha \in \{0.0, 0.25, 0.5, 0.75, 1.0\}$ suffices.
We use $n\!=\!16$ samples per candidate to compute trace vectors, which ensures stable estimates due to the concentration of the squared Frobenius norm.
Because the trace vectors are structured and low-dimensional~($\mathbb{R}^L$, $L\!=\!32$), cosine similarities between candidates naturally concentrate near unity. The proxy's discriminative value lies in the \emph{relative ranking} among candidates, which reliably separates competitive intermediate ratios from dominated extremes.

\subsection{DP Solver Details}
\label{app:dp_details}

We exploit the problem's sequential structure to solve the ILP~(Eq.~\eqref{eq:ilp}) exactly via dynamic programming.
Define $V(l, r)$ as the minimum MOA cost achievable for layers $1, \ldots, l$ using a total of $r$ bits.
The recurrence is:
\begin{equation}
    V(l, r) = \min_{b \in \mathcal{B}} \left[ S_{\text{MOA}}(l, b) + V(l-1, r - b) \right],
\end{equation}
with boundary condition $V(0, 0) = 0$ and $V(0, r) = \infty$ for $r \neq 0$.
The time complexity is $O(L \cdot B_{\text{total}} \cdot |\mathcal{B}|)$, which is negligible~($<\!1$ second for all configurations).

The resulting allocation produces a heterogeneous per-layer bit pattern that interleaves 3-bit, 4-bit, and occasionally 8-bit layers according to the multi-objective sensitivity landscape~(see \Cref{tab:app_allocation}).
For instance, at a 3.5-bit average budget on LLaMA-3-8B, the ILP assigns 16 layers to W3 and 16 layers to W4, with the W4 layers concentrated at the union of PPL-sensitive and reasoning-sensitive positions.

\subsection{Intra-Layer Group-Wise Allocation Details}
\label{app:intra_layer_details}

While the inter-layer ILP determines each layer's \emph{average} bit-width, substantial sensitivity variation exists \emph{within} each layer across different weight groups.
Following the framework of SliM-LLM~\citep{huang2024slim}, we apply GPTQ~\citep{frantar2023optq} for layer-wise quantization and leverage the Optimal Brain Surgeon~(OBS) framework~\citep{hassibi1993obs} to compute per-group salience scores.
For a weight group $\mathcal{G}$ of size $g\!=\!128$ within layer $l$, the OBS salience quantifies the downstream impact of rounding errors,
\begin{equation}
    \text{Salience}(\mathcal{G}) = \sum_{j \in \mathcal{G}} \frac{(\hat{w}_j - w_j)^2}{[\mathbf{H}_l^{-1}]_{jj}},
\end{equation}
where $w_j$ and $\hat{w}_j$ are the original and quantized weight values, and $[\mathbf{H}_l^{-1}]_{jj}$ is the $j$-th diagonal entry of the inverse Hessian.
Groups with higher salience receive higher precision.

For a layer assigned average bit-width $b_l$ by the ILP, each weight group is allocated one of three precision levels ($b_l\!-\!1$, $b_l$, or $b_l\!+\!1$ bits), subject to the constraint that the group-level average matches $b_l$.
Groups are ranked by their OBS salience score; the top-scoring groups receive $b_l\!+\!1$ bits, the bottom-scoring groups receive $b_l\!-\!1$ bits, and the remainder stay at $b_l$.
This two-level architecture enables compression rates that are impossible with layer-level allocation alone.

\section{Extended Experimental Results}
\label{app:extended_results}

\subsection{Sensitivity Heatmaps}
\label{app:sensitivity_heatmaps}

\begin{figure}[t]
\centering
\begin{subfigure}[t]{0.42\columnwidth}
    \centering
    \includegraphics[width=\textwidth]{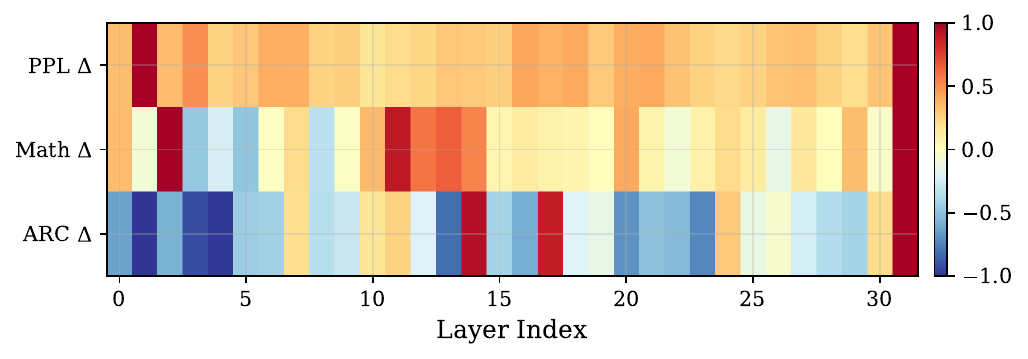}
    \caption{LLaMA-3-8B sensitivity.}
    \label{fig:sens_llama}
\end{subfigure}
\hspace{1em}
\begin{subfigure}[t]{0.42\columnwidth}
    \centering
    \includegraphics[width=\textwidth]{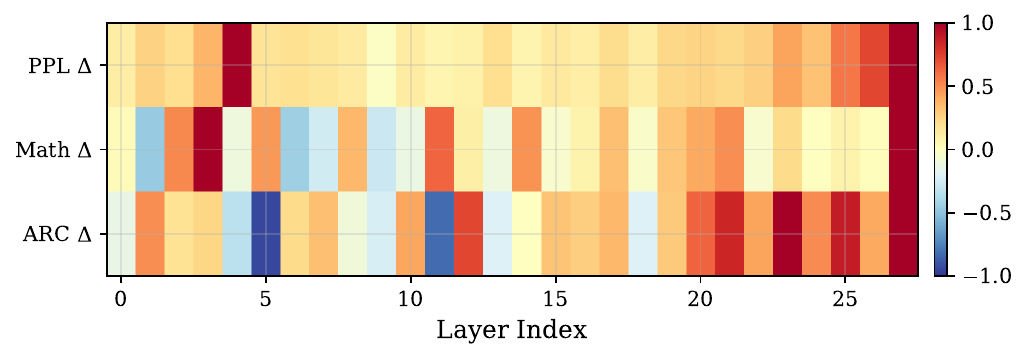}
    \caption{Qwen2.5-7B sensitivity.}
    \label{fig:sens_qwen}
\end{subfigure}
\begin{subfigure}[t]{0.42\columnwidth}
    \centering
    \includegraphics[width=\textwidth]{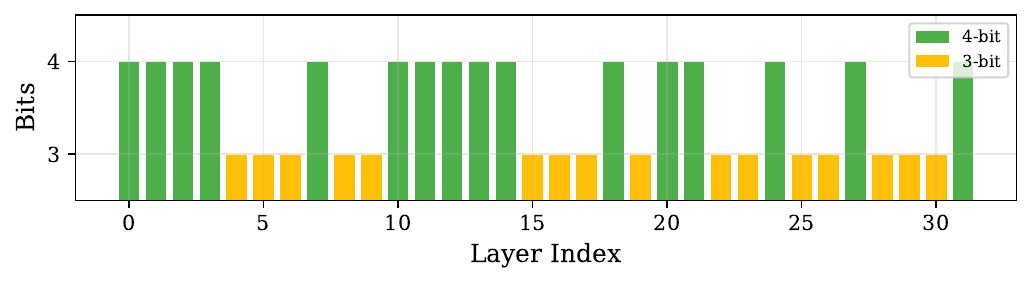}
    \caption{LLaMA-3-8B MOA allocation (b=3.5).}
    \label{fig:alloc_llama}
\end{subfigure}
\hspace{1em}
\begin{subfigure}[t]{0.42\columnwidth}
    \centering
    \includegraphics[width=\textwidth]{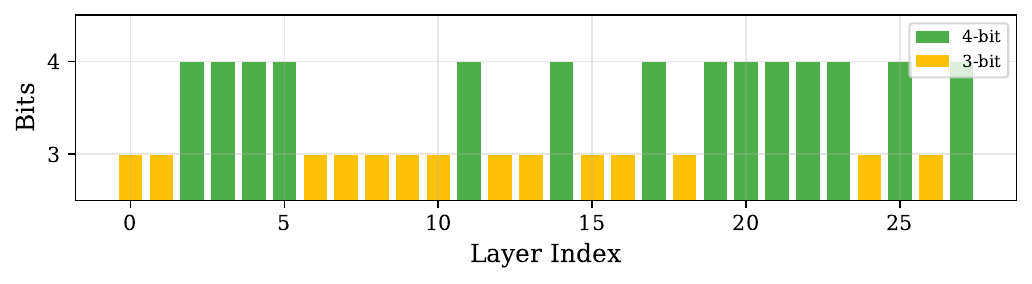}
    \caption{Qwen2.5-7B MOA allocation (b=3.5).}
    \label{fig:alloc_qwen}
\end{subfigure}
\caption{\textbf{Per-layer sensitivity profiles and MOA bit allocation.}
(a)--(b) Layer-wise sensitivity under W3 quantization for three objectives~(symmetric log scale; final layer clipped for visibility).
(c)--(d) Bit-width allocation produced by MOA at b3.5.}
\label{fig:sensitivity_heatmap}
\end{figure}

\Cref{fig:sensitivity_heatmap} visualizes the spatial distribution of per-layer sensitivity across three objectives~(PPL, Math, ARC) and the resulting MOA allocation for both models. PPL-sensitive layers concentrate at the boundary layers, while reasoning-sensitive layers are broadly distributed across the middle layers.

\begin{table*}[t]
\centering
\caption{Comparison with uniform-precision baselines on \textbf{LLaMA-3-8B}. All evaluations on complete test sets. \textbf{Bold}: best in group; \underline{underline}: second best. \method{} at b4.0 achieves the best GSM8K among all W4 methods, and at b3.0 substantially outperforms all W3 baselines. $^\dagger$SpQR's effective bit-widths include grouping metadata and FP16 outlier storage overhead~(see text for details). Intermediate TASA budgets~(b3.25--b3.75) are reported in \Cref{tab:app_budget}.}
\label{tab:main_llama}
\resizebox{\textwidth}{!}{
\begin{tabular}{llccccccccc}
\toprule
\textbf{Method} & \textbf{Bits} & \textbf{ARC-C} & \textbf{HSwag} & \textbf{WinoG} & \textbf{PIQA} & \textbf{ARC-E} & \textbf{BoolQ} & \textbf{GSM8K} & \textbf{PPL}$\downarrow$ & \textbf{Avg.}$\uparrow$ \\
\midrule
FP16 & 16.0 & 53.4 & 79.2 & 72.8 & 79.6 & 80.1 & 81.3 & 49.8 & 7.25 & 70.9 \\
\midrule
RTN & 4.0 & 49.3 & \underline{78.7} & 73.4 & 79.2 & 77.9 & 80.5 & 36.2 & 8.37 & 67.9 \\
GPTQ & 4.0 & 51.0 & 77.6 & 72.4 & 78.8 & \underline{79.0} & 80.2 & 42.3 & 8.35 & 68.8 \\
SpQR$^\dagger$ & 4.63\,(W4) & \textbf{52.8} & \textbf{78.9} & 72.5 & \textbf{79.3} & \textbf{80.3} & \textbf{82.0} & \underline{47.7} & \textbf{7.44} & \textbf{70.5} \\
AWQ & 4.0 & 52.1 & 78.3 & 72.9 & \textbf{79.3} & 78.6 & \underline{80.8} & 43.4 & \underline{7.90} & 69.4 \\
HQQ & 4.0 & 51.1 & 78.2 & \textbf{73.9} & 79.2 & 78.6 & 76.6 & 39.4 & 8.09 & 68.1 \\
\textbf{\method{} b4.0} & \textbf{4.0} & \underline{52.3} & 78.2 & \textbf{73.9} & 79.2 & 79.3 & 79.3 & \textbf{48.0} & 7.92 & \underline{70.0} \\
\midrule
RTN & 3.0 & 32.3 & 53.9 & 59.1 & 68.2 & 54.4 & 63.1 & 0.0 & 47.94 & 47.3 \\
SpQR$^\dagger$ & 4.12\,(W3) & \textbf{49.6} & \textbf{77.9} & 72.5 & \textbf{78.2} & \textbf{77.2} & \textbf{78.3} & \textbf{42.5} & \textbf{7.85} & \textbf{68.0} \\
HQQ & 3.0 & 36.9 & 67.5 & 65.3 & 72.1 & 67.7 & 69.2 & 10.0 & 14.41 & 55.5 \\
OWQ & 3.0 & 42.7 & 71.7 & 68.0 & 74.3 & 70.6 & 70.1 & 19.2 & 13.28 & 59.5 \\
GPTQ & 3.0 & 34.0 & 66.4 & 65.9 & 69.9 & 59.8 & 67.9 & 8.6 & 102.99 & 53.2 \\
\textbf{\method{} b3.0} & \textbf{3.0} & \underline{47.4} & \underline{74.3} & \textbf{72.6} & \underline{77.0} & \underline{75.9} & \underline{77.9} & \underline{39.5} & \underline{10.13} & \underline{66.4} \\
\bottomrule
\end{tabular}
}
\end{table*}

\begin{table*}[t]
\centering
\caption{Comparison with uniform-precision baselines on \textbf{Qwen2.5-7B}. \textbf{Bold}: best in group; \underline{underline}: second best. \method{} at b4.0 ties with HQQ for the best Avg.\ among all W4 methods~(74.7), while at b3.75 retaining 99.1\% of FP16 accuracy. $^\dagger$SpQR effective bit-widths include metadata and outlier overhead~(see \Cref{tab:main_llama} caption). Intermediate TASA budgets~(b3.25--b3.75) are reported in \Cref{tab:app_budget}.}
\label{tab:main_qwen}
\resizebox{\textwidth}{!}{
\begin{tabular}{llccccccccc}
\toprule
\textbf{Method} & \textbf{Bits} & \textbf{ARC-C} & \textbf{HSwag} & \textbf{WinoG} & \textbf{PIQA} & \textbf{ARC-E} & \textbf{BoolQ} & \textbf{GSM8K} & \textbf{PPL}$\downarrow$ & \textbf{Avg.}$\uparrow$ \\
\midrule
FP16 & 16.0 & 51.1 & 78.9 & 72.8 & 78.8 & 80.4 & 84.7 & 83.1 & 8.74 & 75.7 \\
\midrule
RTN & 4.0 & 49.9 & 78.0 & 70.5 & 78.2 & 79.3 & 83.5 & 78.0 & 9.75 & 73.9 \\
GPTQ & 4.0 & 48.5 & 78.1 & 70.9 & \underline{78.6} & 78.1 & 83.8 & \textbf{79.7} & \underline{9.10} & 73.9 \\
AWQ & 4.0 & \underline{51.1} & \underline{78.2} & 71.3 & 78.0 & \underline{79.5} & 82.4 & 78.2 & 9.13 & 74.1 \\
HQQ & 4.0 & 50.9 & \underline{78.2} & \textbf{72.3} & \underline{78.6} & \textbf{80.2} & \underline{84.1} & 78.8 & 9.23 & \textbf{74.7} \\
\textbf{\method{} b4.0} & \textbf{4.0} & \textbf{51.5} & \textbf{78.6} & \underline{71.9} & \textbf{78.7} & 79.2 & \textbf{84.4} & \underline{78.9} & \textbf{9.02} & \textbf{74.7} \\
\midrule
RTN & 3.0 & 34.8 & 49.6 & 56.1 & 63.8 & 54.8 & 71.7 & 0.0 & 106.75 & 47.3 \\
GPTQ & 3.0 & 45.7 & 73.9 & 65.7 & 75.9 & 70.5 & 78.7 & 52.0 & 10.69 & 66.1 \\
SpQR$^\dagger$ & 4.12\,(W3) & \textbf{51.5} & \textbf{78.3} & \textbf{70.6} & \textbf{78.3} & \textbf{80.6} & \textbf{84.2} & \textbf{78.0} & \textbf{9.06} & \textbf{74.5} \\
HQQ & 3.0 & 46.8 & 75.0 & 68.2 & 76.4 & 75.0 & 77.7 & 58.7 & 11.65 & 68.3 \\
OWQ & 3.0 & \underline{50.3} & \underline{75.4} & 66.4 & 77.1 & \underline{79.8} & 79.4 & \underline{60.1} & \underline{10.16} & \underline{69.7} \\
\textbf{\method{} b3.0} & \textbf{3.0} & 47.1 & \underline{75.4} & \underline{68.4} & \underline{78.1} & 76.1 & \underline{82.0} & 59.3 & 9.96 & 69.5 \\
\bottomrule
\end{tabular}
}
\end{table*}

\subsection{Full Baseline Comparison}
\label{app:full_results}

\Cref{tab:main_llama,tab:main_qwen} present the complete per-task comparison between \method{} and all uniform-precision baselines at the W4 and W3 operating points on LLaMA-3-8B and Qwen2.5-7B, respectively.
The full budget sweep across intermediate bit budgets~(b3.25--b3.75) is given in \Cref{tab:app_budget}, and the scaling trends are visualized in \Cref{fig:main_results,fig:pareto}.

\subsection{Full Sensitivity Profiles}
\label{app:sensitivity}

\Cref{tab:app_sens_llama} and \Cref{tab:app_sens_qwen} present the complete per-layer sensitivity values under W3 perturbation for both models, where each value represents the relative change~(\%) in the corresponding metric when only that layer is quantized to 3-bit.

\begin{table*}[t]
\centering
\caption{Complete per-layer sensitivity under W3 perturbation for \textbf{LLaMA-3-8B}. Values are relative change~(\%). The last layer~(L31) exhibits extreme sensitivity across all metrics due to its role as the output projection layer.}
\label{tab:app_sens_llama}
\resizebox{\textwidth}{!}{
\begin{tabular}{l|rrr|l|rrr|l|rrr|l|rrr}
\toprule
& $\Delta$PPL & $\Delta$Math & $\Delta$ARC & & $\Delta$PPL & $\Delta$Math & $\Delta$ARC & & $\Delta$PPL & $\Delta$Math & $\Delta$ARC & & $\Delta$PPL & $\Delta$Math & $\Delta$ARC \\
\midrule
L0 & 1.73 & 0.74 & $-$5.65 & L8 & 1.30 & $-$0.72 & $-$3.05 & L16 & 2.15 & 0.26 & $-$5.22 & L24 & 1.17 & 0.48 & 2.54 \\
L1 & 5.18 & $-$0.19 & $-$8.84 & L9 & 1.40 & $-$0.05 & $-$2.61 & L17 & 1.98 & 0.20 & 7.80 & L25 & 1.31 & 0.27 & $-$1.15 \\
L2 & 1.76 & 2.20 & $-$5.08 & L10 & 0.82 & 0.76 & 1.43 & L18 & 2.12 & 0.14 & $-$1.77 & L26 & 1.59 & $-$0.31 & $-$0.51 \\
L3 & 2.51 & $-$1.04 & $-$8.19 & L11 & 1.13 & 1.97 & 2.25 & L19 & 1.47 & 0.02 & $-$1.26 & L27 & 1.66 & 0.32 & $-$2.07 \\
L4 & 1.32 & $-$0.51 & $-$8.77 & L12 & 1.22 & 1.26 & $-$1.83 & L20 & 2.06 & 0.90 & $-$6.15 & L28 & 1.31 & 0.03 & $-$3.25 \\
L5 & 1.55 & $-$1.08 & $-$3.95 & L13 & 1.54 & 1.42 & $-$7.23 & L21 & 2.14 & 0.16 & $-$4.47 & L29 & 1.03 & 0.73 & $-$3.72 \\
L6 & 2.04 & $-$0.03 & $-$3.80 & L14 & 1.49 & 1.17 & 8.23 & L22 & 1.62 & $-$0.18 & $-$4.70 & L30 & 1.59 & $-$0.08 & 1.81 \\
L7 & 2.01 & 0.47 & 1.77 & L15 & 1.42 & 0.13 & $-$3.69 & L23 & 1.32 & 0.20 & $-$6.43 & L31 & 130.82 & 108.93 & 208.79 \\
\bottomrule
\end{tabular}
}
\end{table*}

\begin{table*}[t]
\centering
\caption{Complete per-layer sensitivity under W3 perturbation for \textbf{Qwen2.5-7B}. L27 is the output projection and shows extreme sensitivity.}
\label{tab:app_sens_qwen}
\resizebox{\textwidth}{!}{
\begin{tabular}{l|rrr|l|rrr|l|rrr|l|rrr}
\toprule
& $\Delta$PPL & $\Delta$Math & $\Delta$ARC & & $\Delta$PPL & $\Delta$Math & $\Delta$ARC & & $\Delta$PPL & $\Delta$Math & $\Delta$ARC & & $\Delta$PPL & $\Delta$Math & $\Delta$ARC \\
\midrule
L0 & 0.70 & 0.28 & $-$2.53 & L7 & 1.00 & $-$3.17 & 5.80 & L14 & 0.42 & 6.04 & $-$0.05 & L21 & 1.39 & 6.25 & 15.16 \\
L1 & 1.60 & $-$5.87 & 8.90 & L8 & 0.80 & 4.52 & $-$1.80 & L15 & 0.88 & $-$0.67 & 5.62 & L22 & 1.66 & $-$0.62 & 7.61 \\
L2 & 1.22 & 6.40 & 3.19 & L9 & $-$0.14 & $-$3.49 & $-$4.15 & L16 & 0.64 & 0.90 & 4.89 & L23 & 2.67 & 2.69 & 18.02 \\
L3 & 2.25 & 12.60 & 4.30 & L10 & 0.75 & $-$1.60 & 7.52 & L17 & 1.32 & 4.10 & 6.37 & L24 & 1.99 & $-$0.06 & 9.10 \\
L4 & 6.25 & $-$1.31 & $-$5.97 & L11 & 0.37 & 7.81 & $-$14.80 & L18 & 0.71 & $-$0.43 & $-$3.70 & L25 & 3.52 & 0.90 & 15.90 \\
L5 & 1.04 & 5.91 & $-$16.96 & L12 & 0.55 & 1.28 & 13.05 & L19 & 1.45 & 3.75 & 5.16 & L26 & 4.53 & 0.12 & 7.36 \\
L6 & 1.20 & $-$5.52 & 3.89 & L13 & 1.24 & $-$1.43 & $-$3.78 & L20 & 1.55 & 5.19 & 11.36 & L27 & 575.63 & 295.69 & 1774.06 \\
\bottomrule
\end{tabular}
}
\end{table*}

Several patterns emerge.
First, the last layer of each model~(L31 for LLaMA, L27 for Qwen) exhibits extreme sensitivity across all metrics, with PPL increases of 130\% and 575\% respectively.
This layer is universally protected at the highest available bit-width~(W8 when budget allows) by all allocation strategies.
Second, excluding the output layer, the PPL sensitivity for LLaMA ranges from 0.82\%~(L10) to 5.18\%~(L1), a 6.3$\times$ dynamic range, while Math sensitivity ranges from $-1.08$\%~(L5) to 2.20\%~(L2), with many layers showing \emph{negative} sensitivity~(quantization improves math performance).
This negative sensitivity phenomenon is consistent with the regularization hypothesis discussed in \Cref{subsec:alignment_diversity}.

\subsection{Cross-Task Sensitivity Correlation}
\label{app:correlation}

We compute the full Kendall rank correlation matrix for the per-layer sensitivity vectors reported in \Cref{tab:kendall_tau} and additionally report the Spearman rank correlation~(\Cref{tab:app_spearman}) for completeness.

\begin{table}[t]
\centering
\caption{Spearman rank correlation $\rho$ between per-layer sensitivity vectors~(W3). Consistent with the Kendall $\tau$ analysis: PPL is uncorrelated with reasoning.}
\label{tab:app_spearman}
\begin{tabular}{l|ccc|ccc}
\toprule
& \multicolumn{3}{c|}{\textbf{LLaMA-3}} & \multicolumn{3}{c}{\textbf{Qwen2.5}} \\
& PPL & Math & ARC & PPL & Math & ARC \\
\midrule
PPL & 1.00 & $-$0.10 & $-$0.28 & 1.00 & 0.03 & 0.53 \\
Math & & 1.00 & 0.29 & & 1.00 & 0.10 \\
ARC & & & 1.00 & & & 1.00 \\
\bottomrule
\end{tabular}
\end{table}

\subsection{Cross-Task Overlap Analysis}
\label{app:overlap_analysis}

A set-theoretic analysis of the top-$K$ most sensitive layers corroborates the rank correlation results discussed in the main text.

On LLaMA-3-8B, the top-8 PPL-sensitive layers are $\{1, 3, 6, 16, 18, 20, 21, 31\}$.
In contrast, the top-8 math-sensitive layers are $\{2, 10, 11, 12, 13, 14, 20, 31\}$, and the top-8 ARC-sensitive layers are $\{7, 10, 11, 14, 17, 24, 30, 31\}$.
As shown in \Cref{tab:overlap}, PPL and Math share only 2 out of 8 layers~(25\%), while PPL and ARC share merely 1 out of 8~(12.5\%).
By contrast, Math and ARC share 4 out of 8 layers~(50\%), strongly suggesting the existence of a shared \emph{reasoning subnetwork} that is functionally distinct from the layers governing perplexity.

\begin{table}[t]
\centering
\caption{Cross-task overlap of top-$K$ sensitive layers under W3 quantization. Random expectation: $K^2/L$.}
\label{tab:overlap}
\begin{tabular}{l|ccc|ccc}
\toprule
& \multicolumn{3}{c|}{\textbf{LLaMA-3 (top-8/32)}} & \multicolumn{3}{c}{\textbf{Qwen2.5 (top-7/28)}} \\
& PPL & Math & ARC & PPL & Math & ARC \\
\midrule
PPL & 8 & 2 & 1 & 7 & 2 & 4 \\
Math & 2 & 8 & 4 & 2 & 7 & 2 \\
ARC & 1 & 4 & 8 & 4 & 2 & 7 \\
\midrule
Random & \multicolumn{3}{c|}{$8^2/32 = 2.0$} & \multicolumn{3}{c}{$7^2/28 = 1.75$} \\
\bottomrule
\end{tabular}
\end{table}

On Qwen2.5-7B~(top-7 of 28 layers), a similar pattern emerges, with PPL$\cap$Math = 2/7~(29\%) and PPL$\cap$ARC = 4/7~(57\%).
The higher PPL--ARC overlap on Qwen reflects this model's stronger integration of scientific knowledge into its language modeling objective during pre-training.

The random expectation for top-$K$ overlap under uniform random selection is $K^2/L$~(2.0 for LLaMA-3, 1.75 for Qwen2.5).
The PPL--Math overlap is consistently at or near this random baseline, indicating that PPL sensitivity is a poor predictor of which layers are critical for mathematical reasoning.
In contrast, the Math--ARC overlap significantly exceeds random expectation, indicating that reasoning tasks share a common set of critical layers that traditional PPL-guided allocation would systematically fail to protect.

\subsection{Auto-Calibration Results}
\label{app:auto_calib_results}

\begin{table}[t]
\centering
\caption{Auto-calibration results. $\alpha^*$ denotes the optimal wiki ratio identified by gradient trace alignment. 
}
\label{tab:auto_calib_results}
\begin{tabular}{ll|ccccc|c}
\toprule
\textbf{Model} & \textbf{Target} & \multicolumn{5}{c|}{Global $\cos$ at $\alpha_{\text{wiki}} =$} & $\alpha^*$ \\
& & 0.0 & 0.25 & 0.50 & 0.75 & 1.0 & \\
\midrule
LLaMA-3 & GSM8K & \textbf{.9998} & .9992 & .9958 & .9878 & .9765 & 0.0 \\
LLaMA-3 & ARC & .9945 & .9983 & \textbf{.9986} & .9953 & .9885 & 0.5 \\
LLaMA-3 & Multi & .9977 & \textbf{.9997} & .9986 & .9934 & .9849 & 0.25 \\
Qwen2.5 & GSM8K & .9999 & \textbf{1.000} & .9996 & .9990 & .9979 & 0.25 \\
Qwen2.5 & ARC & .9927 & .9945 & .9963 & .9977 & \textbf{.9988} & 1.0 \\
\bottomrule
\end{tabular}
\end{table}

\Cref{tab:auto_calib_results} presents the auto-calibration results across five configurations.
Different models and tasks demand different calibration compositions, confirming that no single recipe is universally optimal.

\subsection{Mix Ratio Per-Task Visualization}
\label{app:mix_bar}

\begin{figure}[t]
\centering
\begin{subfigure}[t]{0.42\columnwidth}
    \centering
    \includegraphics[width=\textwidth]{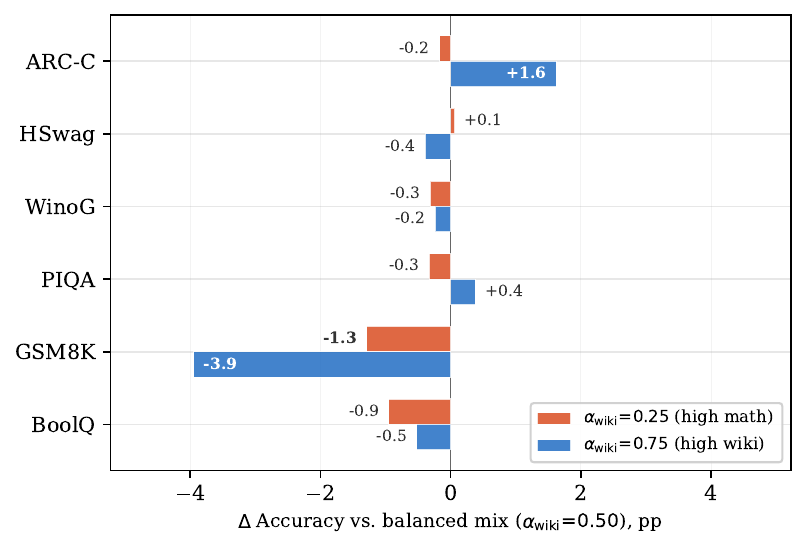}
    \caption{LLaMA-3-8B (b=3.5).}
    \label{fig:mix_llama}
\end{subfigure}
\hspace{1em}
\begin{subfigure}[t]{0.42\columnwidth}
    \centering
    \includegraphics[width=\textwidth]{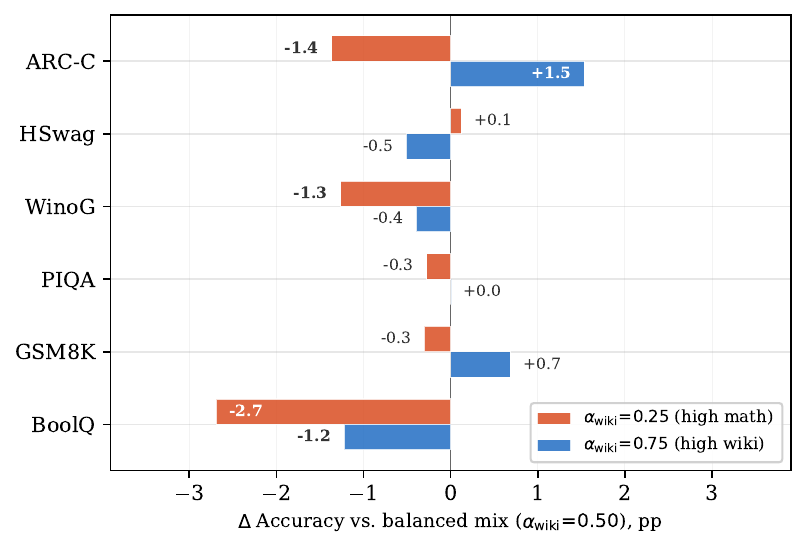}
    \caption{Qwen2.5-7B (b=3.5).}
    \label{fig:mix_qwen}
\end{subfigure}
\caption{
\textbf{Per-task accuracy shifts under extreme calibration mix ratios.} Bars show absolute accuracy changes~(percentage points) relative to the balanced mix~($\alpha_{\text{wiki}}\!=\!0.50$). The direction and magnitude of the shifts are model-dependent, showing the need for the calibration selection.
}
\label{fig:mix_bar}
\end{figure}

\Cref{fig:mix_bar} visualizes the per-task accuracy shifts of two extreme ratios relative to the balanced mix~($\alpha_{\text{wiki}}\!=\!0.50$). Deviating in either direction introduces task-specific trade-offs whose pattern is model-dependent. On LLaMA-3, a high-wiki ratio sharply hurts GSM8K~($-3.9$ pp) while improving ARC-C~($+1.6$ pp). On Qwen2.5, the same ratio instead benefits GSM8K~($+0.7$ pp) but degrades BoolQ and WinoGrande.

\subsection{Auto-Calibration Full Landscapes}
\label{app:auto_calib_full}

\Cref{tab:app_auto_calib_full} presents the complete cosine similarity landscape for all five auto-calibration configurations.
In addition to the global cosine, we report the mean per-layer cosine and the minimum per-layer cosine~(the ``bottleneck'' layer).

\begin{table*}[t]
\centering
\caption{Auto-calibration full landscapes. For each configuration, five candidate mixing ratios are evaluated. Global $\cos$, mean per-layer $\cos$, and minimum per-layer $\cos$~(with the bottleneck layer index) are reported.}
\label{tab:app_auto_calib_full}
\resizebox{\textwidth}{!}{
\begin{tabular}{ll|c|ccc|ccc|ccc|ccc|ccc}
\toprule
\textbf{Model} & \textbf{Target} & & \multicolumn{3}{c|}{$\alpha_{\text{wiki}}\!=\!0.0$} & \multicolumn{3}{c|}{$\alpha_{\text{wiki}}\!=\!0.25$} & \multicolumn{3}{c|}{$\alpha_{\text{wiki}}\!=\!0.50$} & \multicolumn{3}{c|}{$\alpha_{\text{wiki}}\!=\!0.75$} & \multicolumn{3}{c}{$\alpha_{\text{wiki}}\!=\!1.0$} \\
& & $\alpha^*$ & glob & mean & min & glob & mean & min & glob & mean & min & glob & mean & min & glob & mean & min \\
\midrule
LLaMA-3 & GSM8K & 0.0 & \textbf{.9998} & .9999 & .999\textsuperscript{L1} & .9992 & .9996 & .996\textsuperscript{L1} & .9958 & .9985 & .986\textsuperscript{L1} & .9878 & .9959 & .963\textsuperscript{L1} & .9765 & .9920 & .926\textsuperscript{L1} \\
LLaMA-3 & ARC & 0.5 & .9945 & .9976 & .985\textsuperscript{L0} & .9983 & .9986 & .984\textsuperscript{L0} & \textbf{.9986} & .9986 & .982\textsuperscript{L0} & .9953 & .9975 & .972\textsuperscript{L1} & .9885 & .9950 & .943\textsuperscript{L1} \\
LLaMA-3 & Multi & 0.25 & .9977 & .9991 & .995\textsuperscript{L0} & \textbf{.9997} & .9996 & .994\textsuperscript{L0} & .9986 & .9992 & .991\textsuperscript{L0} & .9934 & .9975 & .974\textsuperscript{L1} & .9849 & .9945 & .943\textsuperscript{L1} \\
Qwen2.5 & GSM8K & 0.25 & .9999 & .9998 & .9998\textsuperscript{L0} & \textbf{1.000} & .9997 & .997\textsuperscript{L0} & .9996 & .9987 & .984\textsuperscript{L0} & .9990 & .9968 & .958\textsuperscript{L0} & .9979 & .9936 & .911\textsuperscript{L0} \\
Qwen2.5 & ARC & 1.0 & .9927 & .9947 & .922\textsuperscript{L0} & .9945 & .9956 & .930\textsuperscript{L0} & .9963 & .9960 & .933\textsuperscript{L0} & .9977 & .9956 & .927\textsuperscript{L0} & \textbf{.9988} & .9941 & .902\textsuperscript{L0} \\
\bottomrule
\end{tabular}
}
\end{table*}

A consistent pattern emerges: the ``bottleneck'' layer is always L0 or L1, the input embedding layers, which exhibit the largest sensitivity to the calibration data distribution.
The global cosine metric used for auto-calibration aggregates information from all layers, effectively weighting each layer by its activation energy, and is dominated by these high-sensitivity boundary layers.

\subsection{Three-Way Calibration Search}
\label{app:tri_calib}

In addition to the binary~(wiki vs.\ math) calibration search, we conducted a three-way search over wiki, math, and ARC data for LLaMA-3 targeting GSM8K.
\Cref{tab:app_tri_calib} presents the results for 15 candidate three-way mixing ratios.

\begin{table}[t]
\centering
\caption{Three-way calibration search~(LLaMA-3 $\to$ GSM8K). The optimal three-way mix~($\alpha^*$: w=0, m=0.75, a=0.25) achieves slightly higher cosine than the best binary mix, but the improvement is marginal.}
\label{tab:app_tri_calib}
\begin{tabular}{ccc|cc}
\toprule
Wiki & Math & ARC & Global $\cos$ & Min-layer $\cos$ \\
\midrule
0.00 & 0.00 & 1.00 & .9967 & .986\textsuperscript{L0} \\
0.00 & 0.25 & 0.75 & .9983 & .991\textsuperscript{L0} \\
0.00 & 0.50 & 0.50 & .9993 & .996\textsuperscript{L0} \\
0.00 & 0.75 & 0.25 & \textbf{.9998} & .999\textsuperscript{L0} \\
0.00 & 1.00 & 0.00 & .9998 & .999\textsuperscript{L1} \\
0.25 & 0.00 & 0.75 & .9949 & .983\textsuperscript{L1} \\
0.25 & 0.25 & 0.50 & .9967 & .989\textsuperscript{L1} \\
0.25 & 0.50 & 0.25 & .9979 & .995\textsuperscript{L1} \\
0.25 & 0.75 & 0.00 & .9992 & .996\textsuperscript{L1} \\
0.50 & 0.00 & 0.50 & .9916 & .973\textsuperscript{L1} \\
0.50 & 0.25 & 0.25 & .9936 & .982\textsuperscript{L1} \\
0.50 & 0.50 & 0.00 & .9958 & .986\textsuperscript{L1} \\
0.75 & 0.00 & 0.25 & .9841 & .958\textsuperscript{L1} \\
0.75 & 0.25 & 0.00 & .9878 & .963\textsuperscript{L1} \\
1.00 & 0.00 & 0.00 & .9765 & .926\textsuperscript{L1} \\
\bottomrule
\end{tabular}
\end{table}

The three-way search confirms that the optimal GSM8K calibration is dominated by math data, with ARC data providing a marginal additional alignment boost~($\cos$: 0.99985 for m=0.75/a=0.25 vs.\ 0.99978 for m=1.0/a=0.0).
The practical difference between binary and three-way calibration is negligible, suggesting that the simpler binary search suffices for most applications.

\subsection{Per-Layer Allocation Maps}
\label{app:allocation}

\Cref{tab:app_allocation} shows the MOA allocation patterns for selected configurations.
At b3.5, the ILP produces a 50/50 split between W3 and W4 layers on both models, but the specific layer assignments differ, reflecting the different sensitivity landscapes.

\begin{table}[t]
\centering
\caption{MOA layer allocation patterns~(b3.5). W4 layers are those where MOA sensitivity is highest, protecting both PPL-critical and reasoning-critical layers.}
\label{tab:app_allocation}
\begin{tabular}{l|l}
\toprule
\textbf{Config} & \textbf{W4 layer indices} \\
\midrule
LLaMA-3 b3.5 & 0, 1, 2, 3, 7, 10, 11, 12, 13, \\
& 14, 18, 20, 21, 24, 27, 31 \\
\midrule
Qwen b3.5 & 2, 3, 4, 5, 11, 14, 17, 19, 20, \\
& 21, 22, 23, 25, 27 \\
\bottomrule
\end{tabular}
\end{table}

On LLaMA-3, the W4 layers include both PPL-critical layers~(L0--3, L31) and reasoning-critical layers~(L10--14, L20--21), demonstrating that MOA successfully protects both types of sensitive layers simultaneously.
On Qwen, the W4 layers are more concentrated in the deeper layers~(L17--27), reflecting this model's different sensitivity distribution.

\begin{table}[ht]
\centering
\caption{Calibration seed stability~(3 seeds, MOA + mixed calibration). Only the calibration data sampling order varies; the layer allocation is deterministic. Values are mean$\,\pm\,$std across seeds.}
\label{tab:seed_stability}
\resizebox{\columnwidth}{!}{
\begin{tabular}{lccccccccc}
\toprule
\textbf{Config} & \textbf{ARC-C} & \textbf{HSwag} & \textbf{WinoG} & \textbf{PIQA} & \textbf{ARC-E} & \textbf{BoolQ} & \textbf{GSM8K} & \textbf{PPL}$\downarrow$ & \textbf{Avg.}$\uparrow$ \\
\midrule
LLaMA-3 b3.5 & $50.0 {\scriptstyle \pm 0.3}$ & $76.9 {\scriptstyle \pm 0.1}$ & $72.6 {\scriptstyle \pm 0.7}$ & $78.2 {\scriptstyle \pm 0.2}$ & $77.5 {\scriptstyle \pm 0.5}$ & $79.2 {\scriptstyle \pm 0.3}$ & $45.5 {\scriptstyle \pm 0.8}$ & $8.68 {\scriptstyle \pm 0.03}$ & $68.6 {\scriptstyle \pm 0.3}$ \\
Qwen2.5 b3.75 & $52.5 {\scriptstyle \pm 1.2}$ & $77.4 {\scriptstyle \pm 0.1}$ & $71.2 {\scriptstyle \pm 1.1}$ & $78.1 {\scriptstyle \pm 0.0}$ & $79.6 {\scriptstyle \pm 1.0}$ & $83.3 {\scriptstyle \pm 1.0}$ & $78.3 {\scriptstyle \pm 1.7}$ & $9.40 {\scriptstyle \pm 0.02}$ & $74.3 {\scriptstyle \pm 0.5}$ \\
\bottomrule
\end{tabular}
}
\end{table}

\subsection{Calibration Seed Stability}
\label{app:seed_stability}

\Cref{tab:seed_stability} reports the mean and standard deviation of \method{} across three calibration data sampling seeds at representative operating points: LLaMA-3-8B at b3.5 and Qwen2.5-7B at b3.75.
Only the calibration data sampling order varies across seeds; the MOA sensitivity profiling and ILP allocation are deterministic given the model, so the per-layer bit-width map is identical across all seeds.
The aggregate accuracy~(Avg.) exhibits low variance on both models: $\pm 0.3$ on LLaMA and $\pm 0.5$ on Qwen, confirming that TASA's gains are stable under calibration resampling.
Even the worst individual seed exceeds all uniform-precision W3 baselines by a wide margin~(LLaMA: 68.3 vs.\ OWQ 59.5; Qwen: 73.7 vs.\ OWQ 69.7).
GSM8K shows the largest per-task variance~($\pm 0.8$ on LLaMA, $\pm 1.7$ on Qwen), consistent with the inherent evaluation noise of exact-match scoring on mathematical reasoning.

\subsection{Extended Budget Sweep Data}
\label{app:budget_sweep}

\Cref{tab:app_budget} presents the complete per-task accuracy and perplexity results for the budget sweep experiment, covering bit budgets from b2.5 to b4.0 on both LLaMA-3-8B and Qwen2.5-7B.
All configurations use MOA allocation with mixed calibration.
Performance generally degrades smoothly as the budget decreases, with no large anomalous drops at any operating point.
Both models retain over 90\% of their FP16 Avg.\ at b3.0~(66.4/70.9 for LLaMA, 69.5/75.7 for Qwen), confirming that \method{}'s allocation remains effective even under aggressive compression.
The sharpest degradation occurs between b3.0 and b2.5, where the budget forces the majority of layers to W2 precision.

\begin{table}[t]
\centering
\caption{Full budget sweep results~(\method{}, MOA, mixed calibration) across all seven accuracy tasks and WikiText-2 PPL.}
\label{tab:app_budget}
\setlength{\tabcolsep}{2pt}
\begin{tabular}{lccccccccc}
\toprule
\textbf{Config} & \textbf{ARC-C} & \textbf{HSwag} & \textbf{WinoG} & \textbf{PIQA} & \textbf{ARC-E} & \textbf{BoolQ} & \textbf{GSM8K} & \textbf{PPL}$\downarrow$ & \textbf{Avg.} \\
\midrule
\multicolumn{10}{c}{\textit{LLaMA-3-8B}} \\
\midrule
FP16 & 53.4 & 79.2 & 72.8 & 79.6 & 80.1 & 81.3 & 49.8 & 7.25 & 70.9 \\
b4.0 & 52.3 & 78.2 & 73.9 & 79.2 & 79.3 & 79.3 & 48.0 & 7.92 & 70.0 \\
b3.75 & 50.3 & 77.6 & 73.6 & 78.8 & 78.1 & 79.9 & 46.6 & 8.31 & 69.3 \\
b3.5 & 50.0 & 76.9 & 73.6 & 78.6 & 77.7 & 79.6 & 46.2 & 8.72 & 68.9 \\
b3.25 & 49.3 & 76.2 & 72.2 & 78.5 & 78.0 & 79.7 & 42.5 & 9.10 & 68.1 \\
b3.0 & 47.4 & 74.3 & 72.6 & 77.0 & 75.9 & 77.9 & 39.5 & 10.13 & 66.4 \\
b2.5 & 36.9 & 56.5 & 65.5 & 70.0 & 65.4 & 69.3 & 19.5 & 18.03 & 54.7 \\
\midrule
\multicolumn{10}{c}{\textit{Qwen2.5-7B}} \\
\midrule
FP16 & 51.1 & 78.9 & 72.8 & 78.8 & 80.4 & 84.7 & 83.1 & 8.74 & 75.7 \\
b4.0 & 51.5 & 78.6 & 71.9 & 78.7 & 79.2 & 84.4 & 78.9 & 9.02 & 74.7 \\
b3.75 & 53.9 & 77.4 & 72.3 & 78.1 & 80.6 & 84.3 & 78.4 & 9.38 & 75.0 \\
b3.5 & 50.6 & 77.0 & 70.6 & 77.9 & 79.1 & 84.6 & 79.2 & 9.47 & 74.1 \\
b3.25 & 52.6 & 76.2 & 68.7 & 78.0 & 80.7 & 83.7 & 77.5 & 9.72 & 73.9 \\
b3.0 & 47.1 & 75.4 & 68.4 & 78.1 & 76.1 & 82.0 & 59.3 & 9.96 & 69.5 \\
b2.5 & 37.4 & 54.5 & 61.2 & 70.2 & 64.2 & 67.2 & 47.9 & 16.23 & 57.5 \\
\bottomrule
\end{tabular}
\end{table}

\subsection{GSM8K Accuracy--Compression Tradeoff}
\label{app:gsm8k_pareto}

\begin{figure}[t]
\centering
\begin{subfigure}[t]{0.42\columnwidth}
    \centering
    \includegraphics[width=\textwidth]{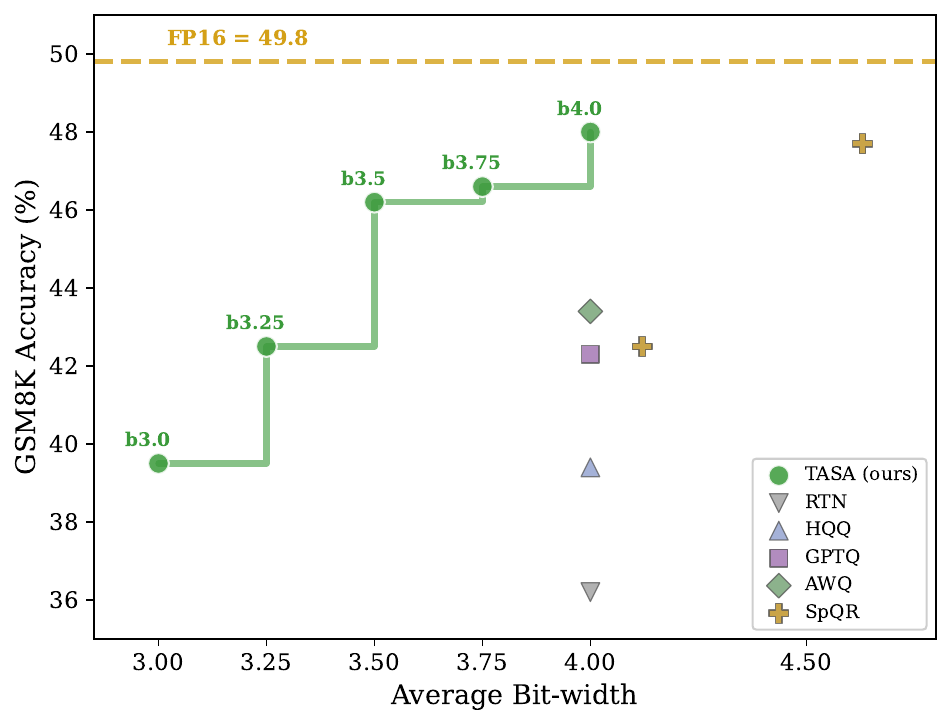}
    \caption{LLaMA-3-8B, GSM8K accuracy.}
    \label{fig:pareto_llama_gsm8k}
\end{subfigure}
\hspace{1em}
\begin{subfigure}[t]{0.42\columnwidth}
    \centering
    \includegraphics[width=\textwidth]{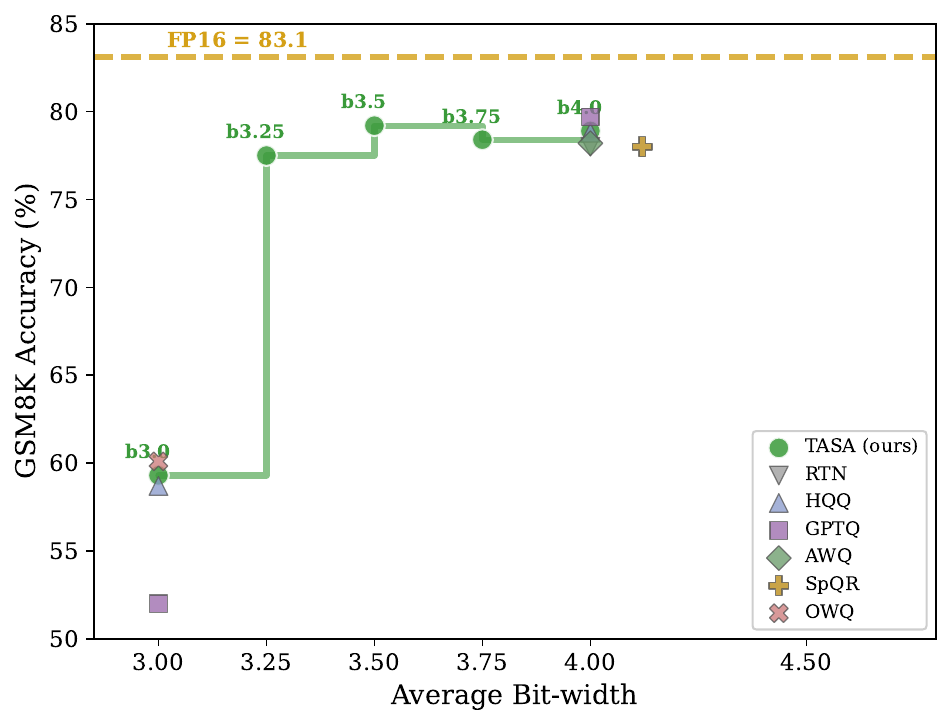}
    \caption{Qwen2.5-7B, GSM8K accuracy.}
    \label{fig:pareto_qwen_gsm8k}
\end{subfigure}
\caption{\textbf{Reasoning accuracy (GSM8K) vs.\ effective bit-width.} The \method{} scaling curve (green) shows smooth progression from 3.0 to 4.0 bits. Muted markers represent uniform baselines at their effective bit-widths (incorporating SpQR's metadata overhead). \method{} outperforms all 3-bit baselines and approaches FP16 reasoning accuracy at 3.5-bit.}
\label{fig:pareto}
\end{figure}

\Cref{fig:pareto} plots GSM8K reasoning accuracy as a function of average bit-width for all \method{} configurations and uniform-precision baselines.
On LLaMA-3-8B, \method{} b3.5 achieves 46.2\% GSM8K, approaching the FP16 ceiling of 49.8\%, while all 4-bit baselines cluster at 36--43\%.
On Qwen2.5-7B, \method{} b3.5 reaches 79.2\% GSM8K, matching the 4-bit baselines~(GPTQ 79.7\%, HQQ 78.8\%).
The \method{} scaling curve demonstrates smooth progression from b3.0 to b4.0, matching or exceeding all baselines at every bit budget.

\subsection{MOA Sensitivity Weight Ablation}
\label{app:beta_ablation}

The MOA aggregation~(Eq.~\eqref{eq:moa}) introduces a single hyperparameter $\beta$ that controls the balance between task-specific protection and PPL preservation.
\Cref{tab:beta_ablation} reports the performance of three $\beta$ values on LLaMA-3-8B at b3.5.

\begin{table*}[t]
\centering
\caption{MOA weight $\beta$ ablation~(LLaMA-3-8B, b3.5, mixed calibration). Performance is stable across $\beta \in [0.5, 0.9]$, with the default $\beta = 0.7$ achieving the highest Avg.}
\label{tab:beta_ablation}
\resizebox{\textwidth}{!}{
\begin{tabular}{cccccccccc}
\toprule
$\beta$ & \textbf{ARC-C} & \textbf{HSwag} & \textbf{WinoG} & \textbf{PIQA} & \textbf{ARC-E} & \textbf{BoolQ} & \textbf{GSM8K} & \textbf{PPL}$\downarrow$ & \textbf{Avg.}$\uparrow$ \\
\midrule
0.5 & 50.2 & 77.0 & 73.2 & 78.1 & 78.0 & 78.2 & 43.8 & 8.70 & 68.4 \\
0.7~(default) & 50.0 & 76.9 & 73.6 & \textbf{78.6} & 77.7 & \textbf{79.6} & \textbf{46.2} & 8.72 & \textbf{68.9} \\
0.9 & \textbf{50.8} & 76.7 & 73.1 & 78.7 & \textbf{78.3} & 79.4 & 44.4 & 8.71 & 68.8 \\
\bottomrule
\end{tabular}
}
\end{table*}

The Avg.\ varies by only 0.5 percentage points across the entire $\beta \in [0.5, 0.9]$ range~(68.4--68.9), showing that MOA is robust to this hyperparameter.
The default $\beta\!=\!0.7$ achieves the highest Avg.~(68.9), confirming that it operates at the performance optimum.
This stability arises because the two sensitivity signals (Math and PPL) produce correlated layer rankings despite their different absolute scales, so the allocation is largely insensitive to the precise mixing weight.

We also compare Z-score normalized MOA~(denoted MOA-norm in \Cref{tab:strategy_ablation}) against the standard MOA formulation.
On LLaMA-3, MOA achieves Avg.$\!=\!$68.9 versus MOA-norm's 68.8~($-$0.1 points), confirming that the raw sensitivity values are already on comparable scales and explicit normalization provides only marginal benefit.

\subsection{Limitations of the Gradient Trace Proxy}
\label{app:auto_calib_lim}

Under single-task targeting, the trace alignment proxy can exhibit over-alignment.
For LLaMA-3 targeting GSM8K alone, the auto-calibration recommends $\alpha^*_{\text{wiki}}\!=\!0.0$~(pure math data), because the trace alignment monotonically increases as task data replaces generic data, as shown in \Cref{tab:auto_calib_results} where $\cos$ rises from $0.9765$ to $0.9998$.
Yet the mix ratio ablation in \Cref{tab:mix_ratio} shows that Avg.\ performance peaks at $\alpha_{\text{wiki}}\!=\!0.50$ and that pure task data is suboptimal, consistent with the Alignment-Diversity Tradeoff of \Cref{subsec:alignment_diversity}.

This discrepancy reveals a fundamental limitation: as a macroscopic energy summary, the trace captures inter-layer energy routing but is blind to the intra-layer eigenspectral collapse illustrated in \Cref{fig:eigenvalue_decay}.
When calibration data is drawn from a single narrow distribution, the per-layer Hessians become spectrally degenerate as predicted by \Cref{prop:diversity_regularization}, but this degeneration is invisible to the trace, which aggregates eigenvalues into a single scalar.
Consequently, the proxy can overfit to alignment at the expense of diversity under extreme single-task configurations.

This limitation is mitigated in practice by two factors.
First, in the multi-task setting~(the primary deployment scenario), the proxy successfully identifies a non-trivial intermediate optimum for LLaMA-3, recommending $\alpha^*\!=\!0.25$ and correctly excluding the extremes.
Second, for Qwen2.5 targeting GSM8K, the proxy recommends $\alpha^*\!=\!0.25$ rather than pure task data, correctly incorporating the model's need for greater diversity.
The single-task over-alignment on LLaMA-3 GSM8K is thus the exception rather than the rule.

\subsection{Activation Alignment Landscape}
\label{app:landscape}

\begin{figure}[t]
\centering
\begin{subfigure}[t]{0.42\columnwidth}
    \centering
    \includegraphics[width=\textwidth]{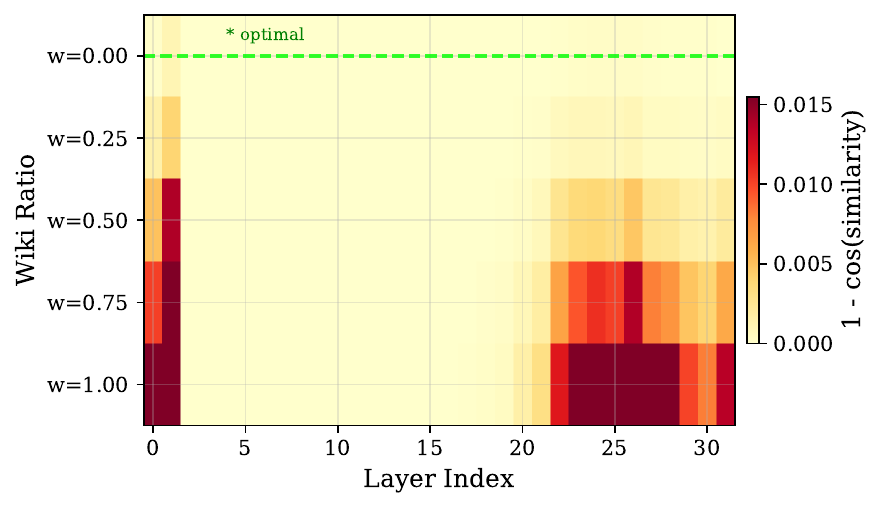}
    \caption{LLaMA-3 $\to$ GSM8K.}
    \label{fig:act_align_gsm8k}
\end{subfigure}
\hspace{1em}
\begin{subfigure}[t]{0.42\columnwidth}
    \centering
    \includegraphics[width=\textwidth]{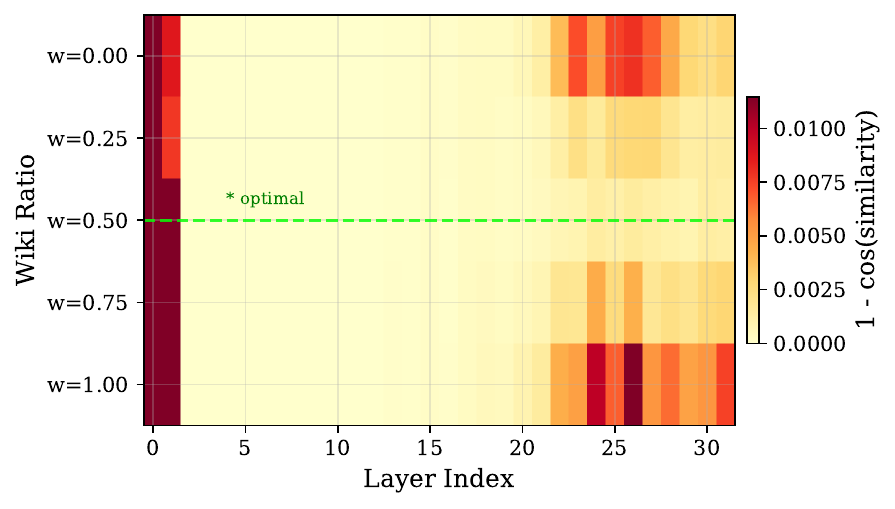}
    \caption{LLaMA-3 $\to$ ARC.}
    \label{fig:act_align_arc}
\end{subfigure}
\caption{\textbf{Per-layer activation divergence heatmap.} Cosine distance ($1-\cos$) between calibration and target-task activations across layers (x-axis) under varying WikiText mix ratios (y-axis). Divergence concentrates at shallow (L0--1) and deep (L22--31) layers, whereas middle layers achieve near-perfect alignment across all configurations. The optimal ratio (marked) is highly task- and model-dependent.}
\label{fig:activation_alignment}
\end{figure}

\Cref{fig:activation_alignment} visualizes the per-layer activation divergence~($1 - \cos$) between different calibration mixtures and the target task, revealing the spatial structure of the alignment landscape.

Two patterns emerge.
First, divergence concentrates at the shallow~(L0--1) and deep~(L22--31) layers, while middle layers~(L2--L20) are nearly perfectly aligned across all calibration configurations.
This reflects the functional specialization of transformer layers: shallow layers handle input embedding and are highly sensitive to the token distribution, while deep layers handle output projection and are similarly distribution-dependent.
Middle layers produce activation patterns that are more universal across data distributions.

Second, different target tasks produce different optimal alignment patterns.
For LLaMA-3 targeting GSM8K, pure math data~($\alpha_{\text{wiki}}\!=\!0$) achieves the best per-layer alignment, while for ARC, a balanced mix~($\alpha_{\text{wiki}}\!=\!0.5$) is optimal.
The auto-calibration search captures these task-specific patterns through the global cosine metric, which aggregates per-layer information into a single scalar that can be efficiently optimized.

\subsection{Eigenspectral Analysis of Calibration Data}
\label{app:eigenspectral}

\Cref{tab:app_eigenspectral} reports the effective rank~(ER) and condition number~($\kappa$) of the per-layer Hessian $\mathbf{H}_l = \mathbf{X}_l^\top \mathbf{X}_l$ under three calibration configurations on LLaMA-3-8B, complementing \Cref{fig:eigenvalue_decay} in \Cref{app:eigenspectral_theory}.

\begin{table}[t]
\centering
\caption{Hessian eigenspectral statistics under different calibration data compositions~(LLaMA-3-8B). ER: effective rank~$\exp(H(\mathbf{p}))$; $\kappa$: condition number $\lambda_{\max}/\lambda_{\min}$.}
\label{tab:app_eigenspectral}
\begin{tabular}{lccc}
\toprule
\textbf{Layer} & \textbf{Calibration} & \textbf{ER} & $\bm{\kappa}$ \\
\midrule
\multirow{3}{*}{L1 (embedding)} & Pure math & 9.5 & 619 \\
& Mixed ($\alpha\!=\!0.5$) & 6.7 & 795 \\
& Pure wiki & 5.0 & 1042 \\
\midrule
\multirow{3}{*}{L12 (middle)} & Pure math & 1.3 & 16654 \\
& Mixed ($\alpha\!=\!0.5$) & 1.4 & 10913 \\
& Pure wiki & 1.5 & 7541 \\
\midrule
\multirow{3}{*}{L22 (reasoning)} & Pure math & 3.3 & 2143 \\
& Mixed ($\alpha\!=\!0.5$) & 4.1 & 1418 \\
& Pure wiki & 5.8 & 950 \\
\midrule
\multirow{3}{*}{L31 (output)} & Pure math & 12.1 & 369 \\
& Mixed ($\alpha\!=\!0.5$) & 15.0 & 300 \\
& Pure wiki & 15.9 & 336 \\
\bottomrule
\end{tabular}
\end{table}

At the reasoning-critical Layer~22, pure math data yields $2.3\times$ higher spectral concentration~($\kappa$: 2143 vs.\ 950) and $1.8\times$ lower effective rank~(3.3 vs.\ 5.8) compared to pure wiki, confirming that task-specific data induces the Hessian degeneracy predicted by \Cref{prop:diversity_regularization}.
Layer~12 exhibits the same trend with even greater magnitude~($\kappa$: 16654 vs.\ 7541), indicating that spectral concentration under task-specific data is a general phenomenon across the middle transformer layers rather than being specific to a single reasoning-critical layer.
The trend reverses at the shallow Layer~1, where general text produces a more concentrated vocabulary distribution~($\kappa$: 1042 vs.\ 619), reflecting the distribution-sensitivity of embedding layers.
Layer~31~(output) shows high effective rank under all calibration conditions~(ER 12--16), consistent with its role as a broad vocabulary projection layer.

\section{Computational Cost Analysis}
\label{app:timing}

\subsection{Stage-by-Stage Breakdown}
\label{app:stage_breakdown}

We analyze the computational complexity of each stage in terms of model forward passes~(hardware-independent) and empirical wall-clock times~(measured on a single A100 GPU for LLaMA-3-8B).

\paragraph{Stage 1: Auto-calibration.}
Computing the trace vector for a single calibration candidate requires one forward pass over $n_{\text{trace}}\!=\!16$ samples. With $|\mathcal{A}|$ candidate mixing ratios plus one target-task reference, the total theoretical cost is $(|\mathcal{A}|+1)$ forward passes. Empirically, this stage completes in $\sim\!4$ minutes.

\paragraph{Stage 2: MOA sensitivity profiling.}
This stage constitutes the primary profiling overhead. For each of the $L$ layers, we quantize it to $|\mathcal{B}|$ candidate bit-widths and evaluate the three objectives over a small calibration subset~($n_{\text{cal}}\!=\!64$ samples per metric). Because the task-specific metrics use conditional cross-entropy loss, all evaluations require only a single forward pass per sample, avoiding the latency of autoregressive generation. The three objectives share the same quantized model state per $(l, b)$ pair, reducing the cost to $L \times |\mathcal{B}| = 96$ quantize-evaluate cycles. Empirically, this multi-objective profiling completes in $\sim\!43$ minutes.

\paragraph{Stage 3: ILP allocation.}
The dynamic programming solution to the inter-layer allocation has negligible complexity and completes in $<\!1$ second. Combining Stages 1--3, the entire \method{}-specific overhead is $\sim\!47$ minutes, a one-time offline cost.

\paragraph{Stage 4: GPTQ backend execution.}
Once the ILP determines the inter-layer allocation, the quantization is executed by a plug-and-play backend. The execution time depends entirely on the chosen backend and is agnostic to \method{}'s allocation. For instance, pairing \method{} with a standard GPTQ backend requires only 1 forward pass per layer for Hessian construction, completing in about $\sim\!10$ minutes~(resulting in a total pipeline time of under 1 hour). Our default implementation uses SliM-LLM's group-wise OBS algorithm to maximize intra-layer mixed-precision quality; while this research backend incurs a higher execution cost, it operates purely as a downstream consumer.

\paragraph{Amortization.}
Stages 1--3 are \emph{one-time costs}. Once the sensitivity vectors and MOA scores are computed, exploring a new global bit budget $B$ requires only re-solving the ILP~(Stage 3) and running the backend quantizer~(Stage 4), bypassing the profiling phase.

\subsection{Full Pipeline Pseudo-Code}
\label{app:pseudocode}

\Cref{alg:tasa} summarizes the complete \method{} pipeline.

\begin{algorithm}[t]
\caption{\method{}: Task-Aware Sensitivity Analysis}
\label{alg:tasa}
\textbf{Input}: Model $\theta$, layers $L$, bit candidates $\mathcal{B}$, budget $B$, general data $\mathcal{D}_g$, task data $\mathcal{D}_t$, candidate ratios $\mathcal{A}$, MOA weight $\beta$ \\
\textbf{Output}: Mixed-precision quantized model $\hat{\theta}$

\begin{algorithmic}[1]
\STATE \textbf{// Stage 1: Auto-Calibration (\Cref{subsec:auto_calib})}
\STATE Compute $\mathbf{h}(\mathcal{D}_t)$ via Eq.~\eqref{eq:trace_vector}
\FOR{$\alpha \in \mathcal{A}$}
    \STATE Mix $\mathcal{D}_{\text{mix}}(\alpha) \leftarrow \alpha \cdot \mathcal{D}_g + (1\!-\!\alpha) \cdot \mathcal{D}_t$
    \STATE Compute $\mathbf{h}(\mathcal{D}_{\text{mix}}(\alpha))$
\ENDFOR
\STATE $\alpha^* \leftarrow \argmax_\alpha \cos(\mathbf{h}(\mathcal{D}_{\text{mix}}(\alpha)),\, \mathbf{h}(\mathcal{D}_t))$
\STATE \textbf{// Stage 2: MOA Sensitivity Profiling (\Cref{subsec:moa_alloc})}
\FOR{$l = 1$ to $L$, \textbf{for} $b \in \mathcal{B}$}
    \STATE $\Delta_l^{(k)}(b) \leftarrow$ Eq.~\eqref{eq:per_layer_sens} using $\mathcal{D}_{\text{mix}}(\alpha^*)$
\ENDFOR
\STATE $S_{\text{MOA}}(l,b) \leftarrow$ Eq.~\eqref{eq:moa} for all $(l, b)$
\STATE \textbf{// Stage 3: ILP Inter-Layer Allocation}
\STATE $\{b_l^*\} \leftarrow$ Solve Eq.~\eqref{eq:ilp} via dynamic programming
\STATE \textbf{// Stage 4: Intra-Layer Quantization (\Cref{subsec:intra_layer})}
\FOR{$l = 1$ to $L$}
    \STATE Apply GPTQ with group-wise mixed-precision at target $b_l^*$
\ENDFOR
\STATE \textbf{return} $\hat{\theta}$
\end{algorithmic}
\end{algorithm}

\subsection{Hardware and Software}
\label{app:hardware}

All experiments are conducted on a single NVIDIA A100-SXM4-40GB GPU using PyTorch 2.1 with CUDA 12.1, and lm-evaluation-harness~\citep{eval-harness} for benchmarking. We note that the A100 does not support native INT4/INT3 compute; all quantized operations are performed with simulated low-precision weights (dequantize-on-the-fly), following standard practice in PTQ literature~\citep{frantar2023optq,lin2024awq}.

\subsection{Theoretical Complexity}
\label{app:complexity}

\Cref{tab:app_complexity} summarizes the forward-pass complexity of each pipeline stage. The profiling stage dominates with $L \times |\mathcal{B}|$ forward passes. However, each profiling pass operates on only $n_{\text{cal}}\!=\!64$ samples, and the per-layer quantize-evaluate cycle touches a single layer while keeping the rest at FP16, making the per-pass cost comparable to a standard inference forward pass.

\begin{table}[h]
\centering
\caption{Theoretical complexity of the \method{} pipeline in terms of model forward passes. $L$: number of layers; $|\mathcal{B}|$: number of bit candidates; $|\mathcal{A}|$: number of mixing ratio candidates.}
\label{tab:app_complexity}
\begin{tabular}{lcc}
\toprule
\textbf{Stage} & \textbf{Forward passes} & \textbf{LLaMA-3 / Qwen2.5} \\
\midrule
Auto-calibration & $|\mathcal{A}|+1$ & 6 / 6 \\
MOA profiling & $L \times |\mathcal{B}|$ & 96 / 84 \\
ILP allocation & 0 & 0 / 0 \\
GPTQ quantization & 1 & 1 / 1 \\
\midrule
\textbf{Total~(first run)} & $|\mathcal{A}| + 1 + L|\mathcal{B}| + 1$ & 103 / 91 \\
\textbf{Incremental~(new $B$)} & 1 & 1 / 1 \\
\bottomrule
\end{tabular}
\end{table}

\paragraph{Comparison with baselines.}
Standard uniform-precision methods~(RTN, GPTQ, AWQ) require a single forward pass for calibration and quantization. \method{}'s overhead is the profiling stage, which is proportional to $L \times |\mathcal{B}|$. This is theoretically comparable to other mixed-precision methods: HAWQ-V2~\citep{dong2020hawqv2} requires $\mathcal{O}(L)$ Hessian eigendecompositions, and SliM-LLM~\citep{huang2024slim} requires $L$ per-layer sensitivity evaluations. The key advantage of \method{} is that the profiling cost is fully amortized across bit budgets.

\subsection{Empirical Wall-Clock Timings}
\label{app:empirical_timing}

\Cref{tab:app_timing_real} details the real-world wall-clock times of the \method{} pipeline. The process is strictly decoupled into the \method{}-specific framework overhead (Stages 1--3) and the backend execution (Stage 4).

\begin{table}[h]
\centering
\caption{Empirical wall-clock time for the \method{} pipeline (Single A100 GPU). The cost is strictly decoupled into TASA-specific framework overhead and backend-dependent execution.}
\label{tab:app_timing_real}
\resizebox{\columnwidth}{!}{
\begin{tabular}{llcc}
\toprule
\textbf{Component} & \textbf{Stage} & \textbf{LLaMA-3-8B} & \textbf{Qwen2.5-7B} \\
\midrule
\multirow{3}{*}{\textbf{\method{} Overhead}}
& Stage 1: Auto-calibration & $\sim 4$ mins & $\sim 4$ mins \\
& Stage 2: MOA Profiling ($n\!=\!64$) & $\sim 43$ mins & $\sim 38$ mins \\
& Stage 3: ILP Allocation & $<\!1$ sec & $<\!1$ sec \\
\cmidrule{2-4}
& \emph{Subtotal (One-time cost)} & \textbf{$\sim 47$ mins} & \textbf{$\sim 42$ mins} \\
\midrule
\multirow{2}{*}{\textbf{Backend Execution}}
& Stage 4: Standard GPTQ (Alternative) & $\sim 10$ mins & $\sim 11$ mins \\
& Stage 4: SliM-LLM GPTQ (Our default) & $\sim 4.9$ hours$^*$ & $\sim 6.1$ hours$^*$ \\
\bottomrule
\multicolumn{4}{l}{\small $^*$ \emph{Includes sequential batch-size-1 operations and per-layer CPU$\leftrightarrow$GPU data transfers; see analysis below.}}
\end{tabular}
}
\end{table}

\paragraph{Analysis of Backend Overhead (Stage 4).}
The \method{} framework is explicitly backend-agnostic. As demonstrated in \Cref{tab:app_timing_real}, if paired with a standard GPTQ backend, the entire \method{} pipeline completes in under 1 hour. Our default experiments employ SliM-LLM~\citep{huang2024slim} as the backend to refine intra-layer precision, which incurs an extended execution duration. Our hardware profiling indicates that this duration is governed by the specific engineering implementation of the underlying research codebase (which utilizes batch-size-1 sequential processing and incurs substantial CPU$\leftrightarrow$GPU data transfer overheads during the OBS salience computation, leading to a GPU compute utilization of $\sim\!10\%$) rather than a mathematical bottleneck. Under time constraints, Stage 4 can be replaced with a standard GPTQ or AWQ backend, preserving \method{}'s inter-layer allocation at lower execution cost.

\section{Baseline Analysis}
\label{app:baseline_analysis}

\subsection{SpQR Effective Bit-Width Calculation}
\label{app:spqr_bits}

SpQR~\citep{dettmers2023spqr} introduces two sources of storage overhead beyond the nominal $b$-bit weight representation, which we detail below.
All numbers are derived from our own reproduction runs~(not the original paper) using the official SpQR codebase with default configuration: group size $g=16$, bilevel quantization with 3-bit scales and 3-bit zeros, and outlier threshold $\sigma_{\text{thr}}=0.2$.

\paragraph{Grouping metadata overhead.}
SpQR uses group size $g=16$, meaning each group of 16 weights shares one scale and one zero-point parameter.
Unlike standard group quantization~(where scales and zeros are stored in FP16), SpQR applies a second level of quantization~(``bilevel'') to compress these group parameters to 3-bit integers.
The 3-bit scale parameters are themselves grouped into second-level groups of $g_2=16$, each sharing an FP16 second-level scale and an FP16 second-level zero-point; the 3-bit zero parameters are handled identically.
The per-weight overhead from a single quantization parameter channel~(e.g., scales) is therefore:
\begin{equation}
\begin{aligned}
b_{\text{channel}} &= \underbrace{\frac{b_{\text{qq}}}{g}}_{\text{3-bit param}} + \underbrace{\frac{2 \times 16}{g \times g_2}}_{\text{FP16 2nd-level scale+zero}} \\
&= \frac{3}{16} + \frac{32}{256} = 0.3125 \text{ bits/weight},
\end{aligned}
\end{equation}
where $b_{\text{qq}}=3$ is the bilevel bit-width, $g=16$ is the primary group size, and $g_2=16$ is the second-level group size.
Since both the scale channel and the zero-point channel incur this overhead, the total grouping metadata cost is:
\begin{equation}
b_{\text{group}} = 2 \times b_{\text{channel}} = 2 \times 0.3125 = 0.625 \text{ bits/weight}.
\label{eq:spqr_group}
\end{equation}

\paragraph{Outlier isolation overhead.}
SpQR detects outlier weights~(those whose removal would cause large reconstruction error) and stores them in uncompressed FP16 precision~(32 bits per value, since both the weight value and its column index must be stored).
The fraction of outlier weights varies by layer; \Cref{tab:app_spqr_outlier} reports the per-layer outlier share from our reproduction logs on LLaMA-3-8B.
The per-weight overhead from outlier isolation is
\begin{equation}
b_{\text{outlier}} = f_{\text{ol}} \times (32 - b),
\label{eq:spqr_outlier}
\end{equation}
where $f_{\text{ol}}$ is the fraction of outlier weights and $b$ is the nominal bit-width.

\begin{table}[t]
\centering
\caption{Per-layer outlier share~(\%) for SpQR on LLaMA-3-8B from our reproduction logs.}
\label{tab:app_spqr_outlier}
\begin{tabular}{cccc|cccc}
\toprule
\textbf{Layer} & \textbf{W3 (\%)} & \textbf{W4 (\%)} & & \textbf{Layer} & \textbf{W3 (\%)} & \textbf{W4 (\%)} & \\
\midrule
L0 & 1.76 & 0.068 & & L16 & 1.71 & 0.032 \\
L1 & 1.72 & 0.043 & & L17 & 1.71 & 0.032 \\
L2 & 1.72 & 0.031 & & L18 & 1.70 & 0.028 \\
L3 & 1.70 & 0.020 & & L19 & 1.69 & 0.027 \\
L4 & 1.71 & 0.026 & & L20 & 1.69 & 0.028 \\
L5 & 1.72 & 0.025 & & L21 & 1.69 & 0.030 \\
L6 & 1.71 & 0.021 & & L22 & 1.68 & 0.025 \\
L7 & 1.72 & 0.022 & & L23 & 1.68 & 0.023 \\
L8 & 1.73 & 0.027 & & L24 & 1.68 & 0.024 \\
L9 & 1.74 & 0.029 & & L25 & 1.68 & 0.026 \\
L10 & 1.73 & 0.028 & & L26 & 1.68 & 0.028 \\
L11 & 1.73 & 0.027 & & L27 & 1.68 & 0.030 \\
L12 & 1.72 & 0.030 & & L28 & 1.68 & 0.030 \\
L13 & 1.72 & 0.031 & & L29 & 1.68 & 0.033 \\
L14 & 1.73 & 0.036 & & L30 & 1.69 & 0.046 \\
L15 & 1.73 & 0.036 & & L31 & 1.71 & 0.051 \\
\midrule
\multicolumn{3}{c}{\textbf{Average}} & & \multicolumn{3}{c}{W3: 1.71\%, W4: 0.031\%} \\
\bottomrule
\end{tabular}
\end{table}

\paragraph{Total effective bit-width.}
Combining both overhead sources, the effective bit-width for SpQR is:
\begin{equation}
b_{\text{eff}} = b + b_{\text{group}} + b_{\text{outlier}} = b + 0.625 + f_{\text{ol}} \times (32 - b).
\end{equation}
For our LLaMA-3-8B reproduction:
\begin{align}
\text{SpQR ``W3'':} \quad b_{\text{eff}} &= 3 + 0.625 + 0.0171 \times 29 \notag \\
&= 4.12 \text{ bits/weight}, \\
\text{SpQR ``W4'':} \quad b_{\text{eff}} &= 4 + 0.625 + 0.0003 \times 28 \notag \\
&= 4.63 \text{ bits/weight}.
\end{align}

This means SpQR ``W3'' uses 37\% more bits than its nominal 3-bit label suggests, and SpQR ``W4'' uses 16\% more.
When comparing at equal effective bit-widths, \method{} b3.5~(3.50 true bits) outperforms SpQR ``W3''~(4.12 effective bits) on Avg.~(0.689 vs.\ 0.680) while using 15\% fewer bits per weight.
\method{}'s reported bit-widths, by contrast, are true average bit-widths with no hidden overhead; inter-layer and intra-layer mixed precision is fully accounted for in the stated budget.

\subsection{Comparison with Task-Specific Calibration (TACQ)}
\label{app:tacq}

TACQ~\citep{xiao2025tacq} takes a different approach to task-aware quantization.
Rather than balancing alignment and diversity at the calibration level, TACQ uses backward-pass gradient attribution to identify ``task circuits,'' the 0.35\% of weight parameters most critical for a specific target task, and preserves them at higher precision while uniformly quantizing the remainder to $b$ bits.
Each \{model, target task, bit-width\} combination requires a separate importance computation involving full backward passes through the model, making the method inherently task-specific.

We reproduce TACQ on both LLaMA-3-8B and Qwen2.5-7B at 2-bit and 3-bit, conditioning on MMLU and GSM8K independently.
\Cref{tab:app_tacq} presents the results alongside \method{} and GPTQ for reference.

\begin{table}[t]
\centering
\caption{Comparison with TACQ~\citep{xiao2025tacq}, a task-specific weight-level mixed-precision method. TACQ uses gradient-based circuit discovery to identify 0.35\% of critical weights per task and preserves them at higher precision; the remaining weights are uniformly quantized to $b$ bits. Each TACQ configuration requires a separate backward-pass importance computation for each \{model, task, bit-width\} combination. ``Calib'' indicates the calibration data distribution. TACQ results use the authors' evaluation protocol (MMLU: 5-shot, last 25\% of test set; GSM8K: 1319 samples with greedy decoding). \method{} results use lm-evaluation-harness (GSM8K: 8-shot CoT, full test set). Due to these protocol differences, cross-method comparisons should be interpreted qualitatively rather than as exact head-to-head numbers.}
\label{tab:app_tacq}
\begin{tabular}{llcccc}
\toprule
\textbf{Method} & \textbf{Bits} & \textbf{Calib} & \textbf{MMLU$^*$} & \textbf{GSM8K$^\dagger$} & \textbf{Avg.} \\
\midrule
\multicolumn{6}{c}{\textit{LLaMA-3-8B}} \\
\midrule
TACQ (MMLU-cond.) & $\sim$2 & MMLU & 49.3 & --- & --- \\
TACQ (MMLU-cond.) & $\sim$3 & MMLU & 63.6 & --- & --- \\
TACQ (GSM8K-cond.) & $\sim$2 & GSM8K & --- & 36.6 & --- \\
TACQ (GSM8K-cond.) & $\sim$3 & GSM8K & --- & 68.3 & --- \\
\midrule
GPTQ & 3.0 & Wiki & --- & 8.6 & 53.2 \\
\method{} b2.5 & 2.5 & Mixed & --- & 19.5 & 54.7 \\
\method{} b3.0 & 3.0 & Mixed & --- & 39.5 & 66.4 \\
\method{} b3.5 & 3.5 & Mixed & --- & 46.2 & 68.9 \\
FP16 & 16.0 & --- & --- & 49.8 & 70.9 \\
\midrule
\multicolumn{6}{c}{\textit{Qwen2.5-7B}} \\
\midrule
TACQ (MMLU-cond.) & $\sim$2 & MMLU & 56.4 & --- & --- \\
TACQ (MMLU-cond.) & $\sim$3 & MMLU & 72.0 & --- & --- \\
TACQ (GSM8K-cond.) & $\sim$2 & GSM8K & --- & 49.1 & --- \\
TACQ (GSM8K-cond.) & $\sim$3 & GSM8K & --- & 79.8 & --- \\
\midrule
GPTQ & 3.0 & Wiki & --- & 52.0 & 66.1 \\
\method{} b2.5 & 2.5 & Mixed & --- & 47.9 & 57.5 \\
\method{} b3.0 & 3.0 & Mixed & --- & 59.3 & 69.5 \\
\method{} b3.5 & 3.5 & Mixed & --- & 79.2 & 74.1 \\
FP16 & 16.0 & --- & --- & 83.1 & 75.7 \\
\bottomrule
\multicolumn{6}{l}{\footnotesize $^*$TACQ protocol: 5-shot, last 25\% of MMLU test set.} \\
\multicolumn{6}{l}{\footnotesize $^\dagger$TACQ protocol: greedy decoding, 1319 samples. \method{}: 8-shot CoT, full test set.} \\
\end{tabular}
\end{table}

Several observations emerge from this comparison.

\paragraph{Task-specific calibration achieves strong single-task performance.}
When calibration and evaluation share the same task distribution, TACQ achieves high scores.
On GSM8K-conditioned 3-bit quantization, TACQ reaches 68.3\% on LLaMA-3 and 79.8\% on Qwen2.5, substantially exceeding \method{} b3.0~(39.5\% and 59.3\% respectively).
This confirms that task-specific calibration has significant potential for maximizing single-task performance, validating the ``alignment'' axis of our Alignment-Diversity Tradeoff framework~(\Cref{subsec:alignment_diversity}).

\paragraph{The generalization cost is substantial.}
TACQ's strong single-task numbers come at a fundamental cost: each target task requires an independent pipeline run, and the resulting quantized model is optimized for that task alone.
The MMLU-conditioned model cannot be evaluated on GSM8K~(and vice versa) without re-running the entire importance computation, which involves full backward passes through the model on task-specific calibration data.
In contrast, \method{} produces a single quantized model from one calibration run that simultaneously achieves competitive performance across all seven evaluation tasks~(Avg.$\!=\!$66.4 at b3.0, 68.9 at b3.5 on LLaMA-3).

\paragraph{Connection to the Alignment-Diversity Tradeoff.}
TACQ can be understood as the extreme alignment endpoint~($\alpha \to 0$) in the framework of \Cref{subsec:alignment_diversity}: it maximizes task alignment by using purely task-specific data for both calibration and importance scoring, at the cost of completely sacrificing calibration diversity.
\method{} occupies the optimal interior point~($\alpha^* \in (0,1)$), trading some single-task optimality for robust cross-task performance.
The gap between TACQ's single-task score and \method{}'s cross-task score quantifies the practical cost of the alignment-diversity balance, a cost that is well justified in deployment scenarios where the model must handle diverse inputs.

\paragraph{Evaluation protocol caveat.}
TACQ uses its own evaluation protocol~(MMLU: 5-shot on the last 25\% of the test set; GSM8K: greedy decoding on 1319 samples), which differs from our lm-evaluation-harness setup~(GSM8K: 8-shot chain-of-thought, full test set).
These protocol differences affect the absolute numbers, so the comparison should be interpreted as qualitative evidence for the alignment-diversity framework rather than a precise head-to-head benchmark.

\section{Theoretical Formalization of the Alignment-Diversity Tradeoff}
\label{app:proofs}

This appendix formalizes the Alignment-Diversity Tradeoff introduced in \Cref{subsec:alignment_diversity}.

\subsection{Diversity Gap Definition}
\label{app:diversity_gap_def}

\begin{definition}[Diversity Gap]
\label{def:diversity_gap}
Given a target task $\mathcal{T}$ and a calibration distribution $\mathcal{D}(\alpha)$ parameterized by the generic-data mixing ratio $\alpha$, the \emph{diversity gap} at mixing ratio $\alpha$ is:
\begin{equation}
    \delta(\alpha) = \cos\!\big(\mathbf{h}(\mathcal{D}(\alpha)),\, \mathbf{h}(\mathcal{D}_{\mathcal{T}})\big) - \frac{\text{Accuracy}(\alpha)}{\max_{\alpha'} \text{Accuracy}(\alpha')},
    \label{eq:diversity_gap}
\end{equation}
where $\mathbf{h}(\cdot)$ denotes the per-layer activation energy trace~(Eq.~\eqref{eq:trace_vector}).
The diversity gap measures the portion of quantization quality that cannot be explained by activation alignment alone, attributable to the regularizing effect of distributional diversity.
\end{definition}

\begin{proposition}[Informal: Diversity as Hessian Regularization]
\label{prop:diversity_regularization}
Quantization under a spectrally degenerate Hessian preserves the narrow task-relevant subspace but distorts the model's broader representational manifold, degrading both general capabilities and the general-purpose features that reasoning itself depends upon.
General-domain data acts as a structural regularizer that lifts the suppressed trailing eigenvalues and maintains a well-conditioned Hessian, preserving the model's generalization structure.
\end{proposition}

\begin{theorem}[Existence of Optimal Mixed Calibration]
\label{thm:optimal_mix}
Let $f(\alpha) = \operatorname{tr}(\mathbf{H}_{\text{test}}\,\mathbf{H}_\alpha^{-1})$.
Assume
\begin{enumerate}[label=(\roman*),nosep]
    \item \label{a:gen} $\mathbf{H}_{\text{gen}} \succ 0$~(generic data is strictly positive definite);
    \item \label{a:task} $\mathbf{H}_{\text{task}}$ is ill-conditioned with $\lambda_{\min}(\mathbf{H}_{\text{task}}) = \epsilon \to 0$, and $\mathbf{H}_{\text{test}}$ has non-zero projection on the trailing eigenspace;
    \item \label{a:align} Task data provides stronger principal alignment than generic data.
\end{enumerate}
Then $f(\alpha)$ is strictly convex on $(0,1]$, $\lim_{\alpha \to 0} f(\alpha) \to \infty$, and $f'(1) > 0$.
Consequently, $f$ attains a unique global minimum at some $\alpha^* \in (0, 1)$.
\end{theorem}

\Cref{thm:optimal_mix} establishes that the optimal calibration is simultaneously model-specific and task-specific, ruling out any fixed heuristic recipe. The full proof is provided in \Cref{app:proof_theorem1}.

\subsection{OBS Derivation of the Trace Objective}
\label{app:obs_derivation}

We ground the Alignment-Diversity Tradeoff under the Optimal Brain Surgeon~(OBS) framework~\citep{hassibi1993obs} and its modern extension to LLM quantization~\citep{frantar2023optq}.
Following standard assumptions in post-training quantization, the raw rounding error $\epsilon$ for each weight column is modeled as uncorrelated zero-mean noise with variance $\sigma_q^2$ determined by the bit-width.
Under OBS, however, quantizing each weight triggers a compensating update to the remaining weights via the inverse calibration Hessian $\mathbf{H}_\alpha^{-1}$, so the effective weight perturbation $\Delta\mathbf{W}$ has covariance proportional to $\sigma_q^2\,\mathbf{H}_\alpha^{-1}$ rather than $\sigma_q^2\,\mathbf{I}$.
Consequently, the expected loss increase on a test distribution $\mathcal{D}_{\text{test}}$ becomes:
\begin{equation}
    \mathbb{E}_{\text{test}}[\Delta \mathcal{L}] \;\approx\; \frac{\sigma_q^2}{2}\,\operatorname{tr}\!\bigl(\mathbf{H}_{\text{test}}\,\mathbf{H}_\alpha^{-1}\bigr),
    \label{eq:trace_error}
\end{equation}
where $\mathbf{H}_\alpha = \alpha\,\mathbf{H}_{\text{gen}} + (1-\alpha)\,\mathbf{H}_{\text{task}}$ is the calibration Hessian under mixing ratio~$\alpha$, and $\mathbf{H}_{\text{test}} = \mathbf{X}_{\text{test}}^\top\mathbf{X}_{\text{test}}/n$ is the Hessian on the test distribution.

This trace formula admits a spectral decomposition that reveals a classical \emph{bias-variance dilemma}.
Let $\{(\lambda_i, \mathbf{u}_i)\}$ be the eigen-pairs of $\mathbf{H}_\alpha$.
Then:
\begin{equation}
    \operatorname{tr}\!\bigl(\mathbf{H}_{\text{test}}\,\mathbf{H}_\alpha^{-1}\bigr) = \sum_{i=1}^{d} \frac{\mathbf{u}_i^\top \mathbf{H}_{\text{test}}\,\mathbf{u}_i}{\lambda_i(\mathbf{H}_\alpha)}.
    \label{eq:spectral_decomp}
\end{equation}

When $\alpha \to 0$~(pure task data), $\mathbf{H}_\alpha$ tightly aligns with $\mathbf{H}_{\text{test}}$ in the dominant eigendirections, yielding small numerators for the leading terms~(\emph{low approximation error}).
However, task-specific distributions span a narrow subspace, causing $\mathbf{H}_{\text{task}}$ to be severely ill-conditioned: its trailing eigenvalues $\lambda_i \to 0$, and the corresponding terms in Eq.~\eqref{eq:spectral_decomp} explode, amplifying quantization noise~(\emph{high instability error}).
This ill-conditioning assumption is supported by empirical analyses of neural network Hessian spectra~\citep{sagun2018empirical,ghorbani2019investigation}, which show that the eigenvalue density is dominated by a large bulk of near-zero eigenvalues.
Conversely, general-domain data acts as a Tikhonov-like structural regularizer that elevates the trailing eigenvalues and ensures $\mathbf{H}_\alpha$ remains well-conditioned, at the cost of reduced alignment in the dominant task directions.

\subsection{Eigenspectral Evidence}
\label{app:eigenspectral_theory}

\begin{figure*}[t]
\centering
\begin{subfigure}[t]{0.28\textwidth}
    \centering
    \includegraphics[width=\textwidth]{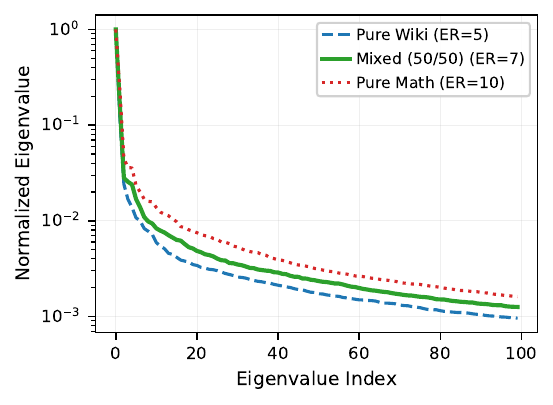}
    \caption{Layer~1 (Shallow).}
    \label{fig:eigen_shallow}
\end{subfigure}
\hspace{1em}
\begin{subfigure}[t]{0.28\textwidth}
    \centering
    \includegraphics[width=\textwidth]{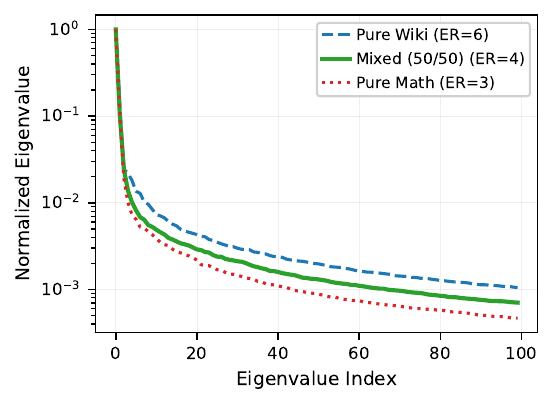}
    \caption{Layer~22 (Reasoning).}
    \label{fig:eigen_reasoning}
\end{subfigure}
\hspace{1em}
\begin{subfigure}[t]{0.28\textwidth}
    \centering
    \includegraphics[width=\textwidth]{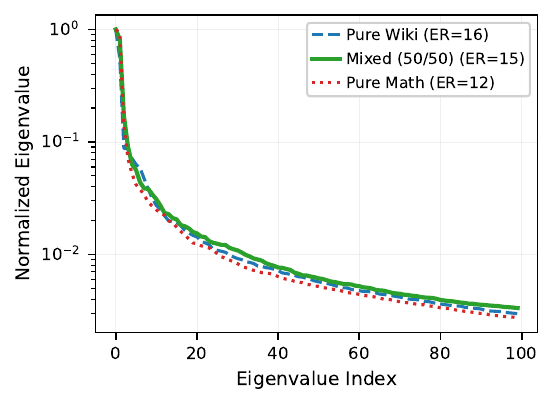}
    \caption{Layer~31 (Output).}
    \label{fig:eigen_output}
\end{subfigure}
\caption{\textbf{Hessian eigenvalue decay across calibration compositions (LLaMA-3-8B).} Each panel plots the top-100 normalized eigenvalues of the layer-wise Hessian ($\mathbf{H}_l = \mathbf{X}_l^\top \mathbf{X}_l$). ER denotes the effective rank. At the reasoning-critical Layer~22, pure math data yields a sharply degenerate spectrum (ER$\!=\!$3) compared to pure Wiki (ER$\!=\!$6), confirming that task-specific data concentrates Hessian energy into fewer directions. Mixed data robustly interpolates these extremes. Conversely, shallow embedding layers (Layer 1) exhibit reversed behavior due to their high sensitivity to token distributions.}
\label{fig:eigenvalue_decay}
\end{figure*}

\Cref{fig:eigenvalue_decay} provides direct evidence for the structural regularization effect of general-domain data by comparing the Hessian eigenspectrum under pure math, mixed, and pure Wiki calibration at three representative layers of LLaMA-3-8B~(exact statistics in \Cref{tab:app_eigenspectral} in \Cref{app:eigenspectral}).

At the reasoning-critical Layer~22, pure math data produces a sharply degenerate spectrum with effective rank~(ER) 3 and condition number $\kappa\!=\!2143$, compared to ER$\!=\!$6 and $\kappa\!=\!950$ under pure Wiki, a $2.3\times$ increase in spectral concentration.
The mixed calibration interpolates between the two~(ER$\!=\!$4, $\kappa\!=\!$1418), confirming that general-domain data acts as the predicted structural regularizer.
At the output Layer~31, the same pattern holds~(pure math ER$\!=\!$12 vs.\ Wiki ER$\!=\!$16).
The trend reverses at the shallow Layer~1~(pure math ER$\!=\!$10 vs.\ Wiki ER$\!=\!$5), reflecting the distribution-sensitivity of embedding layers where general text produces a more concentrated vocabulary distribution.
This layer-dependent behavior motivates a data-composition strategy that balances alignment and diversity \emph{globally} rather than optimizing for any single layer.

\begin{remark}[Diversity as Effective Rank Enhancement]
\label{rem:erank}
The regularizing effect of general-domain data can be quantified through the effective rank~\citep{roy2007effective} $\operatorname{erank}(\mathbf{H}) = \exp(-\sum_i \bar{\lambda}_i \log \bar{\lambda}_i)$, where $\bar{\lambda}_i = \lambda_i / \sum_j \lambda_j$.
Increasing $\alpha$ smooths the eigenspectrum of $\mathbf{H}_\alpha$ by lifting suppressed trailing eigenvalues, thereby increasing the Shannon entropy of the normalized spectrum and raising the effective rank.
\end{remark}

\subsection{Proof of \texorpdfstring{\Cref{thm:optimal_mix}}{Theorem 1}}
\label{app:proof_theorem1}

We prove that the expected quantization error $f(\alpha) = \operatorname{tr}(\mathbf{H}_{\text{test}}\,\mathbf{H}_\alpha^{-1})$ attains a unique global minimum at a non-trivial interior point $\alpha^* \in (0, 1)$.

\paragraph{Detailed assumptions.}
The three assumptions stated in \Cref{thm:optimal_mix} are motivated as follows.

\paragraph{Assumption~\ref{a:gen}: Full-rank generic data.}
Generic corpora~(e.g., WikiText) exercise a broad range of activation directions, yielding a well-conditioned covariance with $\lambda_{\min}(\mathbf{H}_{\text{gen}}) \geq \mu > 0$.

\paragraph{Assumption~\ref{a:task}: Ill-conditioned task data.}
$\mathbf{H}_{\text{task}}$ has minimum eigenvalue $\lambda_{\min}(\mathbf{H}_{\text{task}}) = \epsilon$ with $\epsilon \ll \mu$.
Let $\mathbf{v}$ be the corresponding eigenvector, with $\mathbf{v}^\top \mathbf{H}_{\text{test}} \mathbf{v} = c > 0$.
This is justified by empirical analyses of neural network Hessian spectra~\citep{sagun2018empirical,ghorbani2019investigation} showing that narrow-domain data exacerbates ill-conditioning.

\paragraph{Assumption~\ref{a:align}: Superior task alignment.}
$\operatorname{tr}(\mathbf{H}_{\text{test}}\,\mathbf{H}_{\text{gen}}^{-1}\,\mathbf{H}_{\text{task}}\,\mathbf{H}_{\text{gen}}^{-1}) > \operatorname{tr}(\mathbf{H}_{\text{test}}\,\mathbf{H}_{\text{gen}}^{-1})$.
This holds whenever the task and test distributions share the same domain, so that $\mathbf{H}_{\text{task}}$ captures the principal activation directions of $\mathbf{H}_{\text{test}}$ more faithfully than $\mathbf{H}_{\text{gen}}$ does.

\paragraph{Step 1: Strict convexity of $f(\alpha)$.}
For any fixed $\mathbf{A} \succ 0$, the matrix function $\mathbf{X} \mapsto \operatorname{tr}(\mathbf{A}\,\mathbf{X}^{-1})$ is strictly convex over the positive definite cone, as established in classical convex optimization~\citep{boyd2004convex} and matrix analysis~\citep{bhatia2009positive}.
Since $\mathbf{H}_\alpha$ is a positive affine function of $\alpha$ and is strictly positive definite for all $\alpha \in (0, 1]$~(guaranteed by \Cref{a:gen}), the composite function $f(\alpha) = \operatorname{tr}(\mathbf{H}_{\text{test}}\,\mathbf{H}_\alpha^{-1})$ is strictly convex on $(0, 1]$.

\paragraph{Step 2: Boundary behavior as $\alpha \to 0$.}
As $\alpha \to 0$, we have $\mathbf{H}_\alpha \to \mathbf{H}_{\text{task}}$.
Consider the spectral contribution of the trailing eigenvector $\mathbf{v}$ from \Cref{a:task}:
\begin{equation}
    f(\alpha) \;\geq\; \frac{\mathbf{v}^\top \mathbf{H}_{\text{test}}\,\mathbf{v}}{\lambda_{\min}(\mathbf{H}_\alpha)} \;=\; \frac{c}{\alpha\,\mathbf{v}^\top\mathbf{H}_{\text{gen}}\,\mathbf{v} + (1-\alpha)\epsilon}.
\end{equation}
As $\alpha \to 0$, the denominator approaches $\epsilon \to 0$, so $f(\alpha) \to c/\epsilon \to \infty$.
This proves that pure task data leads to error explosion due to matrix inversion instability, and therefore $\alpha^* \neq 0$.

\paragraph{Step 3: Derivative at $\alpha = 1$.}
The first derivative of $f$ is obtained by the standard matrix calculus identity $\frac{d}{d\alpha}\operatorname{tr}(\mathbf{A}\,\mathbf{X}^{-1}) = -\operatorname{tr}(\mathbf{A}\,\mathbf{X}^{-1}\frac{d\mathbf{X}}{d\alpha}\mathbf{X}^{-1})$:
\begin{equation}
    f'(\alpha) = -\operatorname{tr}\!\bigl(\mathbf{H}_{\text{test}}\,\mathbf{H}_\alpha^{-1}\,(\mathbf{H}_{\text{gen}} - \mathbf{H}_{\text{task}})\,\mathbf{H}_\alpha^{-1}\bigr).
\end{equation}
Evaluating at $\alpha = 1$~(where $\mathbf{H}_{\alpha=1} = \mathbf{H}_{\text{gen}}$):
\begin{align}
    f'(1) &= -\operatorname{tr}\!\bigl(\mathbf{H}_{\text{test}}\,\mathbf{H}_{\text{gen}}^{-1}\,(\mathbf{H}_{\text{gen}} - \mathbf{H}_{\text{task}})\,\mathbf{H}_{\text{gen}}^{-1}\bigr) \notag \\
    &= -\operatorname{tr}\!\bigl(\mathbf{H}_{\text{test}}\,\mathbf{H}_{\text{gen}}^{-1}\bigr) + \operatorname{tr}\!\bigl(\mathbf{H}_{\text{test}}\,\mathbf{H}_{\text{gen}}^{-1}\,\mathbf{H}_{\text{task}}\,\mathbf{H}_{\text{gen}}^{-1}\bigr).
    \label{eq:fprime1}
\end{align}
Under \Cref{a:align}, the second term strictly exceeds the first, giving $f'(1) > 0$.
This means that decreasing $\alpha$ from~1~(i.e., injecting task data into the calibration set) strictly decreases the expected quantization error, so $\alpha^* \neq 1$.

\paragraph{Conclusion.}
Since $f(\alpha)$ is strictly convex on $(0, 1]$, $\lim_{\alpha \to 0} f(\alpha) = +\infty$, and $f'(1) > 0$~(the function is still increasing at the right boundary), the unique global minimum of $f$ must lie in the open interval $(0, 1)$.
This concludes the proof that an optimal mixed calibration strictly outperforms both pure generic and pure task calibration. \hfill$\blacksquare$

\paragraph{Discussion.}
\Cref{a:align} deserves further justification.
When the task and test distributions share the same domain~(e.g., both involve mathematical reasoning), $\mathbf{H}_{\text{task}}$ concentrates energy along the same principal directions as $\mathbf{H}_{\text{test}}$.
The condition $\operatorname{tr}(\mathbf{H}_{\text{test}}\,\mathbf{H}_{\text{gen}}^{-1}\,\mathbf{H}_{\text{task}}\,\mathbf{H}_{\text{gen}}^{-1}) > \operatorname{tr}(\mathbf{H}_{\text{test}}\,\mathbf{H}_{\text{gen}}^{-1})$ states that $\mathbf{H}_{\text{task}}$, when projected through the inverse generic Hessian, amplifies the test-relevant directions more than the identity would.
In the limiting case where $\mathbf{H}_{\text{task}} = \mathbf{H}_{\text{test}}$~(perfect domain match), this reduces to $\operatorname{tr}(\mathbf{M}^2) > \operatorname{tr}(\mathbf{M})$ for $\mathbf{M} = \mathbf{H}_{\text{test}}\,\mathbf{H}_{\text{gen}}^{-1} \succ 0$, which holds whenever $\mathbf{M}$ has any eigenvalue exceeding~1, i.e., whenever the task distribution has at least one direction of higher activation energy than the generic distribution.
This is a very mild condition, satisfied in all practical scenarios where the target task exhibits any distributional specialization.

\end{document}